\documentclass[preprint,12pt,authoryear]{elsarticle}
\usepackage{lineno,hyperref}
\usepackage{ragged2e}
\usepackage[margin=1in]{geometry}
\usepackage{graphicx,subfigure}
\usepackage{multirow}
\usepackage{epstopdf}
\usepackage{setspace}
\usepackage{scalerel}
\usepackage{bbold}
\usepackage{float}
\usepackage{amsmath, amssymb, ulem}
\usepackage{rotating}
\usepackage{morefloats}
\usepackage{stackengine}
\usepackage{comment}
\usepackage{caption}
\usepackage{amsmath,amssymb,amsfonts}
\usepackage{graphicx}
\usepackage{booktabs, rotating, multicol, multirow}
\usepackage{algorithm}
\usepackage{algorithmic}
\usepackage{hyperref}
\usepackage{enumitem}
\usepackage{xcolor}
\hypersetup{
    colorlinks=true,
    linkcolor=blue,
    citecolor=blue,
    urlcolor=blue
}

\usepackage{natbib}
\usepackage{booktabs, rotating, multicol, multirow}
\setcitestyle{round}
\usepackage{fancybox,color}
\usepackage{booktabs}
\usepackage[utf8]{inputenc}
\usepackage[english]{babel}

\usepackage{multirow}

\usepackage{adjustbox}
\usepackage{amsmath,amssymb,amsthm}
\usepackage{thmtools}
\declaretheoremstyle[
spaceabove=6pt, spacebelow=6pt,
headfont=\normalfont\bfseries,
notefont=\mdseries, notebraces={(}{)},
bodyfont=\normalfont,
postheadspace=0.6em,
headpunct=:
]{mystyle}

\usepackage{graphicx}
\usepackage{cleveref}
\crefname{hyp}{hypothesis}{hypotheses}
\Crefname{hyp}{Hypothesis}{Hypotheses}
\usepackage{rotating,tikz}
\usepackage{subfigure}

\usepackage{xcolor}

\usepackage{subcaption}

\makeatletter 
\renewcommand\@biblabel[1]{} 
\makeatother

\journal{Expert Systems with Applications}

\usepackage[margin=1in]{geometry}
\usepackage{amsmath,amsthm,amssymb,amsfonts,graphicx,dsfont}

\usepackage{booktabs,parskip}
\usepackage{epsfig}

\usepackage{url}
\usepackage{bm}
\usepackage{rotating}
\usepackage{listings}
\usepackage{verbatim}
\usepackage{xcolor}
\usepackage[title]{appendix}
\usepackage{multirow}

\usepackage{longtable}
\usepackage{adjustbox}
\usepackage{subfigure}
\hypersetup{colorlinks,%
citecolor=black,%
filecolor=black,%
linkcolor=black,%
urlcolor=black
}

\usepackage{setspace} % for \onehalfspacing

\newcommand{\eval}[2][\right]{\relax
	\ifx#1\right\relax \left.\fi#2#1\rvert}

\newtheoremstyle{break}
  {\topsep}{\topsep}%
  {\itshape}{}%
  {\bfseries}{}%
  {\newline}{}%
\theoremstyle{break}

\newtheorem{proposition}{Proposition}
\theoremstyle{definition}

\newtheorem{exampleemph}[proposition]{Example}   % upshape

\usepackage{changepage}   % for the adjustwidth environment
\makeatletter
\newcommand*{\rom}[1]{\expandafter\@slowromancap\romannumeral #1@}
\makeatother

\begin{document}

% \usepackage[margin=10pt,labelfont=bf]{caption}
% \captionsetup[table]{labelsep=none}
% %\usepackage{subcaption}
% \usepackage{float}
% %\usepackage[capposition=top]{floatrow}
% \newcommand\tcaptab[1]{\captionsetup{position=top, font=normalsize, labelfont=bf, textfont=normalfont, justification=centering, margin=0mm, aboveskip=1mm, belowskip=0mm, labelsep=colon, singlelinecheck=false}\caption{#1}}
% \newcommand\bnotetab[1]{\captionsetup{position=bottom, font=footnotesize,  textfont=normalfont, margin=1mm, skip=2mm, justification=justified, singlelinecheck=false}\caption*{#1}}
% \newcommand\tcapfig[1]{\captionsetup{position=top, font=normalsize, labelfont=bf, textfont=normalfont, justification=centering, margin=0mm, aboveskip=2mm, belowskip=0mm, labelsep=colon, singlelinecheck=false}\caption{#1}}
% \newcommand\bnotefig[1]{\captionsetup{position=bottom, font=footnotesize,  textfont=normalfont, margin=1mm, skip=2mm, justification=justified, singlelinecheck=false}\caption*{#1}}
% \newcommand\subcap[1]{\captionsetup{position=bottom, font=small, labelfont=bf, textfont=normalfont, justification=RaggedRight, margin=0mm, aboveskip=-5mm, belowskip=0mm, labelsep=space, singlelinecheck=false}\caption{#1}}
% \newcommand\subcaptab[1]{\captionsetup{position=bottom, font=small, labelfont=bf, textfont=normalfont, justification=centering, margin=0mm, aboveskip=4mm, belowskip=0mm, labelsep=space, singlelinecheck=false}\caption{#1}}

% PUT ALL FIGURES AND TABLES AT THE END OF DOCUMENT FOR NOW
%\usepackage[nomarkers]{endfloat}

%\journal{Expert Systems with Applications}

\begin{frontmatter}

\title{Topology-Aware Gaussian Graph Repair for Robust Graph Neural Networks
}

\author[mymainaddress]{Anubha Goel\corref{mycorrespondingauthor}}
\cortext[mycorrespondingauthor]{Corresponding author. Tel.: +358 503201908.}
\ead{anubha.goel@tuni.fi}		
		\author[mymainaddress]{Juho Kanniainen}
		\ead{juho.kanniainen@tuni.fi}
		
		\address[mymainaddress]{Computing Science/Financial Computing and Data Analytics Group, Tampere University, Tampere, 33720, Finland}
		
%\end{frontmatter}

%\begin{frontmatter}
		
\newpage
\begin{abstract}

Graph neural networks have achieved strong performance on graph-structured data, but their effectiveness depends heavily on the quality of the observed graph. In real applications, graph topology is often imperfect: noisy edges may connect unrelated nodes, while missing edges may prevent useful information from being propagated. Existing robust graph learning methods mainly address this problem by removing suspicious edges or by learning a new graph structure during training. However, edge removal alone cannot recover missing connections, and graph structure learning may introduce additional optimization complexity. In this paper, we propose Topology-Aware Gaussian Repair (TAGR), a simple graph repair framework for robust message passing in graph neural networks. Instead of learning a dense adjacency matrix, TAGR constructs a sparse feature-neighborhood graph using an adaptive Gaussian kernel and combines it with a topology-aware residual correction of the observed graph. The Gaussian repair component introduces auxiliary edges between feature-similar nodes, while the residual correction preserves and reweights the original topology according to local feature and structural consistency. The repaired graph can be used directly with standard graph neural networks without changing their architectures. Extensive experiments on benchmark citation networks show that TAGR improves the robustness of GNNs under both noisy-edge and missing-edge settings. The analysis further show that Gaussian feature-neighborhood repair provides the main robustness gain, while topology-aware residual correction improves stability when the observed graph is incomplete. These results suggest that effective graph robustness can be achieved through lightweight sparse graph repair rather than dense graph structure learning.

\end{abstract}
\end{frontmatter}
\section{Introduction}
\label{sec:introduction}

Graph neural networks (GNNs) have become a widely used framework for learning from graph-structured data. Their success is largely due to the message-passing mechanism, in which each node updates its representation by aggregating information from its neighbors. Through this process, GNNs combine node attributes with relational structure and have achieved strong performance in node classification, link prediction, recommendation, and other graph learning tasks \cite{kipf2017gcn, hamilton2017graphsage, velickovic2018gat}. Although different GNN architectures define neighborhood aggregation in different ways, they all rely on the observed graph to specify where information is allowed to flow. GCN propagates features through a normalized adjacency matrix, GraphSAGE aggregates sampled local neighborhoods, and GAT learns attention weights over observed neighbors. Therefore, the quality of the input graph directly affects the quality of the learned representations.

In practice, however, the observed graph is rarely a perfect description of task-relevant relations. Real-world graphs may contain spurious edges, missing edges, or links that are structurally valid but weakly related to the downstream prediction task. In citation networks, for example, two papers may discuss similar topics without citing each other, while a citation can also connect papers from different research areas. In social and interaction networks, edges may reflect incidental contact, noisy behavior, or external factors that do not correspond to semantic similarity. When such imperfect graphs are used for message passing, noisy edges can propagate irrelevant information, while missing edges can prevent useful information from reaching related nodes. This makes structural uncertainty a central challenge for robust graph representation learning.

A large body of research has therefore studied how to improve the graph used by GNNs. One direction focuses on graph denoising and sparsification. DropEdge randomly removes edges during training to reduce overfitting and over-smoothing \cite{rong2020dropedge}. NeuralSparse learns a supervised edge-selection mechanism to retain task-relevant neighborhoods \cite{zheng2020neuralsparse}. PTDNet learns to drop task-irrelevant edges through a parameterized topological denoising network \cite{luo2021ptdnet}. These methods show that reducing unreliable message passing can improve robustness. However, edge removal is only one side of the problem. If the observed graph is incomplete, removing edges cannot restore missing propagation paths and may further weaken useful connectivity.

Another direction is graph structure learning, where the graph is refined or reconstructed jointly with GNN training. Methods such as LDS, Pro-GNN, GAug, and related approaches learn or modify graph structures using edge distributions, structural priors, feature similarity, or neural edge prediction \cite{franceschi2019lds, jin2020prognn, zhao2021gaug}. More recent robust graph learning methods further incorporate self-supervision, multi-view learning, or structural representations to improve graph quality under noise and sparse supervision \cite{dai2022towards, liu2026graph}. These approaches are expressive and can add, delete, or reweight edges according to the learning objective. At the same time, many of them require additional trainable graph modules, dense pairwise edge scoring, bilevel optimization, or task-dependent structure learning objectives. Such complexity can increase computational cost, reduce scalability, and make it harder to separate the effect of graph repair from the effect of the downstream GNN architecture.

These observations suggest that robust GNN learning should not be viewed only as edge denoising or full graph structure learning. A useful intermediate goal is \emph{graph repair}: the construction of a message-passing graph that preserves reliable observed topology, compensates for missing connections, and reduces the influence of unreliable edges, while remaining sparse and compatible with standard GNN backbones. Such a repair mechanism should be simple enough to avoid dense graph learning, but expressive enough to improve both noisy and incomplete graphs.

Motivated by this perspective, we propose \emph{Topology-Aware Gaussian Repair} (TAGR), a lightweight graph repair framework for improving the robustness of GNN message passing under structural uncertainty. TAGR constructs a repaired sparse adjacency matrix before GNN training and then uses this graph directly with a standard message-passing architecture. The method combines two complementary components. First, TAGR builds an adaptive Gaussian feature-neighborhood graph. For each node, it identifies a small set of feature-neighbor candidates and assigns locally scaled Gaussian weights, allowing feature-consistent nodes to exchange information even when no observed edge connects them. This component is designed to restore missing semantic neighborhoods without constructing a dense adjacency matrix. Second, TAGR applies a topology-aware residual reweighting mechanism to the observed graph. Instead of treating all original edges as equally reliable, TAGR adjusts their weights using local feature agreement and structural consistency, including neighborhood overlap, common-neighbor information, clustering behavior, and degree imbalance. The final repaired graph combines feature-based auxiliary connectivity with a refined version of the observed topology.

TAGR has several desirable properties. It is sparse: the repair is restricted to feature-neighborhood candidates and observed edges, avoiding dense all-pairs graph learning. It is modular: the repaired graph can be used with different GNN backbones without modifying their neural architectures. It is interpretable: the repair is driven by feature proximity and local topological evidence rather than an opaque graph generator. Most importantly, TAGR addresses two complementary forms of structural corruption. The Gaussian feature-neighborhood graph helps recover missing propagation paths, while topology-aware residual reweighting stabilizes the contribution of the observed graph when its edges are noisy or incomplete.

We evaluate TAGR in the setting of semi-supervised node classification, where structural corruption directly affects neighborhood aggregation. The experiments are conducted under a unified protocol on benchmark citation networks with clean graphs, random edge addition, and random edge deletion. We compare TAGR with standard GNN backbones, Gaussian-only repair, learned graph repair, and graph structure learning baselines. In addition to representative corruption levels, we evaluate robustness curves over a wide range of perturbation ratios and conduct analysis to isolate the effects of feature-neighborhood repair and topology-aware residual reweighting.

The results show that TAGR improves the robustness of standard GNNs across both noisy-edge and missing-edge settings. Under a fixed GCN backbone, TAGR consistently improves over vanilla GCN and provides stable performance across full edge-addition and edge-deletion robustness curves. The study further confirms that Gaussian feature-neighborhood repair is a strong source of robustness, while topology-aware residual reweighting provides additional stabilization, especially when the observed graph becomes incomplete. These findings indicate that robust graph learning does not always require dense graph structure learning or additional graph-generation modules; a sparse and topology-aware repair operator can provide reliable message-passing structure while remaining compatible with existing GNN architectures.

The main contributions of this work are summarized as follows:
\begin{itemize}
    \item We formulate robust GNN learning from the perspective of \emph{graph repair}, targeting both spurious and missing edges in the message-passing topology rather than focusing only on edge removal or dense graph reconstruction.
    
    \item We propose TAGR, a lightweight topology-aware Gaussian graph repair framework that constructs a sparse repaired adjacency matrix before GNN training and can be used directly with standard message-passing architectures.
    
    \item We introduce an adaptive Gaussian feature-neighborhood repair component that restores feature-consistent propagation paths while avoiding dense pairwise adjacency learning.
    
    \item We design a topology-aware residual reweighting mechanism that refines observed edges using local feature and structural evidence, preserving useful topology while reducing reliance on unreliable connections.
    
    \item We provide a unified robustness evaluation under clean, edge-addition, and edge-deletion settings, showing that TAGR improves GNN stability and that its gains arise from complementary feature-based repair and topology-aware residual correction.
\end{itemize}

\section{Related Work}
\label{sec:related_work}

\subsection{Graph Neural Networks and Structural Reliability}
\label{subsec:related_gnns}

Graph neural networks (GNNs) extend neural representation learning to graph-structured data by propagating and aggregating information over graph neighborhoods. Most GNNs follow a message-passing paradigm, in which each node updates its representation by combining its own features with information received from neighboring nodes. Early graph convolutional models can be broadly categorized into spectral and spatial approaches. Spectral methods define convolution through graph signal processing, while spatial methods directly aggregate information from local neighborhoods. GCN simplifies spectral graph convolution through a first-order approximation and performs efficient propagation using the normalized adjacency matrix~\citep{kipf2017gcn}. GraphSAGE generalizes neighborhood aggregation by learning inductive aggregation functions over sampled neighborhoods, making it suitable for large graphs and unseen nodes~\citep{hamilton2017graphsage}. GAT further improves neighborhood aggregation by assigning attention weights to different neighbors instead of treating all connected nodes uniformly~\citep{velickovic2018gat}.

Although these architectures differ in their aggregation mechanisms, they share a common dependence on the observed graph structure. The graph determines which nodes exchange information during message passing and therefore strongly influences the quality of the learned representations. When the observed topology is reliable, message passing allows GNNs to effectively combine feature and structural information. However, real-world graphs are often imperfect. They may contain spurious edges, missing edges, or links that are structurally valid but weakly related to the prediction task. In such cases, GNNs may propagate misleading information through noisy edges or fail to propagate useful information because important connections are absent.

This sensitivity has motivated research on improving the graph used for message passing. Some methods modify the graph through preprocessing or regularization, while others learn a refined graph structure together with the GNN model. The central challenge is to construct a graph that is more reliable for downstream learning without introducing excessive computational or optimization complexity. This challenge is especially important in semi-supervised settings, where limited labeled data makes the model more dependent on the quality of neighborhood propagation.

\subsection{Robust Graph Learning and Graph Structure Repair}
\label{subsec:related_repair}

A major line of work improves GNN robustness by reducing the effect of unreliable edges. Early preprocessing-based defenses modify the graph before training. For example, GCN-Jaccard removes edges between nodes with low feature similarity, while GCN-SVD uses low-rank approximation of the adjacency matrix to reduce structural perturbations~\citep{wu2019sgc, entezari2020all}. These methods are simple and can be effective against certain types of adversarial or noisy edges, but they rely on fixed assumptions about what constitutes an unreliable edge.

Other methods incorporate denoising or sparsification into the learning process. DropEdge randomly removes edges during training to reduce overfitting and alleviate over-smoothing in deep GNNs~\citep{rong2020dropedge}. NeuralSparse learns a supervised sparsification network that selects task-relevant edges and removes uninformative neighborhood connections~\citep{zheng2020neuralsparse}. PTDNet learns a parameterized topological denoising network that drops task-irrelevant edges and uses structural regularization to encourage a cleaner topology~\citep{luo2021ptdnet}. These methods show that suppressing noisy message passing can improve generalization. However, most denoising and sparsification methods are primarily designed to remove or down-weight edges. This is useful when the main difficulty is spurious connectivity, but it does not fully address incomplete topology. If useful propagation paths are missing, edge removal alone cannot recover them and may further reduce graph connectivity.

Graph structure learning provides a more flexible way to refine the message-passing graph. Instead of assuming that the observed adjacency matrix is fixed, graph structure learning methods infer, reconstruct, or reweight edges using node features, structural priors, supervision signals, or learned edge scores. LDS formulates graph learning as a bilevel optimization problem over discrete edge distributions~\citep{franceschi2019lds}. Pro-GNN jointly learns a graph structure and a GNN model by imposing priors such as sparsity, low rank, and feature smoothness~\citep{jin2020prognn}. GRCN revises the graph by predicting missing edges and adjusting edge weights for downstream learning~\citep{yu2020graph}. GAug uses neural edge prediction to promote likely intra-class edges and demote likely inter-class edges for graph data augmentation~\citep{zhao2021gaug}. RS-GNN learns a denoised and densified graph to improve robustness under noisy graphs with sparse labels~\citep{dai2022towards}. More recent approaches further exploit self-supervision, contrastive learning, or structural representations. For example, JNSGSL jointly uses node features and structural features in a dual-channel graph structure learning framework and aligns the two views through contrastive learning~\citep{liu2026graph}.

These methods demonstrate that modifying graph topology can substantially improve the reliability of GNN message passing. At the same time, many graph structure learning approaches require additional trainable graph-learning modules, dense pairwise edge scoring, bilevel optimization, or task-dependent structure-learning objectives. Such designs are expressive, but they may increase computational cost and make the learned graph sensitive to optimization details or validation supervision. This motivates a simpler alternative: instead of learning an entirely new graph, one can repair the observed graph using sparse and interpretable evidence from both node features and local topology.

Feature similarity offers a natural signal for such repair. In attributed graphs, nodes that are close in feature space may share semantic or functional relationships even when no observed edge connects them. Similarity-based graph construction can therefore recover useful propagation paths that are absent from the original topology. However, simple similarity graphs also have limitations. Dense similarity matrices are computationally expensive, global thresholds can be sensitive to feature scale, and binary k-nearest-neighbor graphs ignore the strength of feature similarity. A useful repair mechanism should therefore add feature-consistent connections in a sparse and locally adaptive way, while still preserving reliable information from the observed graph.

TAGR follows this repair perspective. It constructs a sparse adaptive Gaussian feature neighborhood graph to recover feature-consistent propagation paths and combines it with topology-aware residual reweighting of the observed graph. In this way, TAGR occupies an intermediate position between simple preprocessing and full graph structure learning. It is more flexible than using the raw graph or a binary feature-neighborhood graph, but avoids dense adjacency learning and additional graph-generator optimization. This makes it suitable for improving GNN message passing when the observed topology may contain both spurious and missing edges.
\section{Proposed Method}
\label{sec:method}

In this section, we present the proposed Topology-Aware Gaussian Repair (TAGR) framework.
TAGR is designed as a sparse graph repair operator that improves the message-passing structure
used by a downstream GNN. Given an observed graph and node features, TAGR constructs a
repaired weighted adjacency matrix by combining two complementary components: a Gaussian
feature-neighborhood graph that adds missing feature-consistent connections, and a topology-aware
residual reweighting of the observed graph that preserves and refines the original topology. The
repaired graph is then used by a standard GNN backbone without changing the neural architecture
or the training objective.

\subsection{Problem Formulation}
\label{subsec:problem_formulation}

Let $\mathcal{G}=(\mathcal{V},\mathcal{E},\mathbf{X})$ be an observed attributed graph, where
$\mathcal{V}=\{v_1,\ldots,v_n\}$ is the node set, $\mathcal{E}\subseteq \mathcal{V}\times\mathcal{V}$ is the observed edge set, and
\[
    \mathbf{X}=[\mathbf{x}_1,\mathbf{x}_2,\ldots,\mathbf{x}_n]^\top
    \in \mathbb{R}^{n\times d}
\]
is the node feature matrix. The observed graph is represented by an adjacency matrix
\[
    \mathbf{A}\in \mathbb{R}_{\geq 0}^{n\times n},
\]
where $A_{ij}>0$ indicates an observed edge between nodes $v_i$ and $v_j$. In most benchmark settings, $\mathbf{A}$ is binary, but the formulation also allows weighted graphs.

A standard message-passing GNN uses $\mathbf{A}$ to determine which nodes exchange information. This makes the quality of the observed topology central to representation learning. If $\mathbf{A}$ contains spurious edges, message passing can propagate irrelevant or misleading information. If $\mathbf{A}$ is incomplete, semantically related nodes may not exchange information even when their features indicate strong similarity. Therefore, rather than treating the observed graph as fixed and fully reliable, we aim to construct a repaired graph that is more suitable for message passing.

TAGR constructs a repaired adjacency matrix
\[
    \widetilde{\mathbf{A}}=\mathcal{R}(\mathbf{A},\mathbf{X}),
\]
where $\mathcal{R}(\cdot)$ is a deterministic sparse graph repair operator. A downstream GNN is then trained on $(\mathbf{X},\widetilde{\mathbf{A}})$ instead of $(\mathbf{X},\mathbf{A})$. Importantly, TAGR does not introduce a trainable graph generator and does not learn a dense $n\times n$ adjacency matrix. It repairs the message-passing graph before GNN training using node features and local structural evidence.

\paragraph{Adaptive Gaussian feature-neighborhood repair.}
The first component of TAGR repairs missing feature-consistent connectivity. The motivation is that nodes with similar attributes may represent related semantic concepts even when no observed edge connects them. This is particularly important when the graph is incomplete or when edge deletion removes useful propagation paths.

TAGR first computes feature similarity in a normalized feature space. Let $\bar{\mathbf{x}}_i$ denote the row-normalized feature vector of node $v_i$. For sparse bag-of-words features, row normalization reduces the effect of raw feature magnitude and makes similarity depend primarily on feature composition. The cosine similarity between two nodes is
\[
    \rho_{ij}
    =
    \frac{
        \bar{\mathbf{x}}_i^\top \bar{\mathbf{x}}_j
    }{
        \|\bar{\mathbf{x}}_i\|_2\|\bar{\mathbf{x}}_j\|_2+\epsilon
    },
\]
where $\epsilon>0$ is a numerical stability constant. The corresponding cosine distance is
\[
    \delta_{ij}=1-\rho_{ij}.
\]
Self-pairs are excluded by setting $\delta_{ii}=\infty$, which prevents self-loops from being selected as feature-neighborhood repair edges.

To make the Gaussian repair locally adaptive, TAGR assigns each node its own bandwidth. For node $v_i$, let $\delta_{i,(\sigma_k)}$ denote the $\sigma_k$-th smallest value among $\{\delta_{ij}:j\neq i\}$. The local bandwidth is defined as
\[
    \sigma_i=\delta_{i,(\sigma_k)}+\epsilon.
\]
This scale adapts the Gaussian kernel to the local feature density: nodes in dense feature regions use smaller bandwidths, while nodes in sparse regions use larger bandwidths.

For each candidate pair $(i,j)$, TAGR computes the adaptive Gaussian similarity
\[
    K_{ij}
    =
    \exp\left(
        -\frac{\delta_{ij}^{2}}{\sigma_i\sigma_j+\epsilon}
    \right).
\]
This form uses the local bandwidths of both endpoints. Hence, a candidate edge receives a large weight only when the two nodes are close relative to their local feature neighborhoods.

TAGR does not construct a dense feature-similarity graph. For each node $v_i$, it selects only the top $k$ non-neighbor candidates according to $K_{ij}$. Let
\[
    \mathcal{C}_k(i)
    =
    \operatorname{TopK}_{j}\left(K_{ij}\right)
\]
subject to
\[
    j\neq i,\qquad A_{ij}=0,\qquad A_{ji}=0.
\]
The exclusion of existing observed edges ensures that this component is used to add missing feature-neighborhood connections rather than duplicate original graph edges.

The directed Gaussian candidate matrix is defined as
\[
    B_{ij}
    =
    \begin{cases}
    \alpha_g K_{ij}, & j\in \mathcal{C}_k(i),\\
    0, & \text{otherwise},
    \end{cases}
\]
where $\alpha_g>0$ is the Gaussian repair weight. The undirected Gaussian feature-neighborhood repair graph is obtained by symmetrization:
\[
   \mathbf{A}^{\mathrm{GFN}}
    =
    \max(\mathbf{B},\mathbf{B}^{\top}),
\]
where the maximum is taken element-wise. Thus, $\mathbf{A}^{\mathrm{GFN}}$ is a sparse weighted graph that adds at most $O(kn)$ directed candidates before symmetrization and avoids dense all-pairs graph learning.

\paragraph{Topology-aware residual reweighting.}
The Gaussian feature-neighborhood repair graph restores missing feature-consistent edges, but the observed graph may still contain useful relational information. TAGR therefore does not replace the observed topology. Instead, it refines the original graph through a bounded topology-aware residual reweighting mechanism. The goal is to increase the influence of observed edges that are locally consistent and reduce the influence of edges that appear unreliable from feature or structural evidence.

Let $\mathcal{E}_A$ be the edge set of the symmetrized observed graph. For each observed edge $(i,j)\in\mathcal{E}_A$, TAGR computes local edge statistics that describe feature agreement and structural consistency.

First, feature agreement is measured by cosine similarity:
\[
    c_{ij}=\rho_{ij}.
\]

Second, local neighborhood overlap is measured using the Jaccard coefficient. Let
\[
    \mathcal{N}_A(i)=\{u:A_{iu}>0 \ \text{or}\ A_{ui}>0\}
\]
be the one-hop neighborhood of node $v_i$ in the symmetrized observed graph. The Jaccard overlap is
\[
    J_{ij}
    =
    \frac{
        |\mathcal{N}_A(i)\cap \mathcal{N}_A(j)|
    }{
        |\mathcal{N}_A(i)\cup \mathcal{N}_A(j)|+\epsilon
    }.
\]

Third, TAGR uses the number of common neighbors,
\[
    M_{ij}
    =
    |\mathcal{N}_A(i)\cap \mathcal{N}_A(j)|,
\]
with the logarithmic transformation $\log(1+M_{ij})$ to reduce the dominance of large common-neighbor counts.

Fourth, TAGR incorporates local clustering. Let $\kappa_i$ be the clustering coefficient of node $v_i$ in the symmetrized observed graph. The quantity
\[
    \kappa_i+\kappa_j
\]
captures whether the endpoints of an edge lie in locally cohesive regions.

Finally, TAGR includes degree imbalance,
\[
    |d_i-d_j|,
\]
where $d_i=|\mathcal{N}_A(i)|$. Edges connecting nodes with highly different degrees may act as structurally unbalanced shortcuts, especially when noisy edges are present.

Because these statistics have different scales, each edge-level statistic is standardized across the observed edge set. For an edge-statistic vector $\mathbf{u}\in\mathbb{R}^{|\mathcal{E}_A|}$, define
\[
    \operatorname{stdz}(\mathbf{u})
    =
    \frac{\mathbf{u}-\mu(\mathbf{u})}{\operatorname{std}(\mathbf{u})+\epsilon}.
\]
If the standard deviation is numerically zero, the standardized vector is set to zero.

The topology-aware edge score is then computed as
\[
\begin{aligned}
    q_{ij}
    =
    &\; \beta_1 \operatorname{stdz}(c_{ij})
    + \beta_2 \operatorname{stdz}(J_{ij})
    + \beta_3 \operatorname{stdz}(\log(1+M_{ij}))  \\
    &+ \beta_4 \operatorname{stdz}(\kappa_i+\kappa_j)
    - \beta_5 \operatorname{stdz}(|d_i-d_j|),
\end{aligned}
\]
where $\beta_1,\ldots,\beta_5\geq 0$ control the contribution of each local signal. In our implementation, we use
\[
    (\beta_1,\beta_2,\beta_3,\beta_4,\beta_5)
    =
    (1.50,0.50,0.25,0.10,0.10).
\]
The largest weight is assigned to feature agreement because feature consistency is a strong indicator of semantic compatibility in attributed citation graphs. The Jaccard and common-neighbor terms capture local structural agreement, the clustering term rewards locally cohesive regions, and the degree-imbalance term penalizes structurally asymmetric shortcuts.

The combined score is standardized once more:
\[
    s_{ij}=\operatorname{stdz}(q_{ij}).
\]
TAGR maps this standardized score to a bounded residual multiplier:
\[
    r_{ij}
    =
    \operatorname{clip}
    \left(
        1+\lambda\tanh(s_{ij}),
        r_{\min},
        r_{\max}
    \right),
\]
where $\lambda\geq0$ is the residual strength and $(r_{\min},r_{\max})=(0.5,1.5)$. The hyperbolic tangent limits the effect of outlying scores, while clipping prevents excessive amplification or suppression. Thus, this component does not remove observed edges; it gently reweights them according to local feature-topology consistency. We fix these coefficients to avoid introducing many additional validation-tuned hyperparameters.

Let $\mathbf{R}_{\lambda}$ denote the sparse residual multiplier matrix:
\[
    R_{\lambda,ij}
    =
    \begin{cases}
    r_{ij}, & (i,j)\in\mathcal{E}_A,\\
    0, & \text{otherwise}.
    \end{cases}
\]
The topology-aware reweighted observed graph is
\[
    \mathbf{A}^{\mathrm{TAR}}
    =
    \mathbf{A}\odot\mathbf{R}_{\lambda},
\]
where $\odot$ denotes element-wise multiplication.

\paragraph{Final repaired graph.}
The final TAGR adjacency combines the topology-aware reweighted observed graph with the Gaussian feature-neighborhood repair graph:
\[
    \widetilde{\mathbf{A}}
    =
    \mathbf{A}^{\mathrm{TAR}}
    +
    \mathbf{A}^{\mathrm{GFN}}.
\]
Equivalently, TAGR can be written as the sparse graph repair operator
\[
    \mathcal{R}(\mathbf{A},\mathbf{X})
    =
    \underbrace{\mathbf{A}\odot\mathbf{R}_{\lambda}}
    _{\text{topology-aware residual reweighting}}
    +
    \underbrace{\mathcal{G}^{\mathrm{GFN}}_{k,\sigma,\alpha}
    (\mathbf{X};\mathbf{A})}
    _{\text{Gaussian feature-neighborhood repair}},
\]
where $\mathcal{G}^{\mathrm{GFN}}_{k,\sigma,\alpha}(\mathbf{X};\mathbf{A})$ constructs a sparse Gaussian feature-neighborhood repair graph from node features. Here, $k$ controls the number of feature-neighbor candidates added per node, $\sigma$ denotes the locally adaptive Gaussian bandwidth, and $\alpha$ is the Gaussian repair weight. Existing observed edges are excluded from the Gaussian candidate set, so the two terms have complementary roles: $\mathbf{A}\odot\mathbf{R}_{\lambda}$ preserves and refines the observed topology, whereas $\mathcal{G}^{\mathrm{GFN}}_{k,\sigma,\alpha}(\mathbf{X};\mathbf{A})$ restores missing feature-consistent propagation paths.

When $\lambda=0$, the residual multiplier satisfies $r_{ij}=1$ for every observed edge, and topology-aware residual reweighting is disabled. In this case, TAGR reduces to Gaussian feature-neighborhood repair added to the original graph. When the Gaussian repair weight is also set to zero, no auxiliary feature-neighborhood edges are introduced, the repaired graph reduces to the original observed graph, and the model becomes the corresponding base GNN.

The construction above directly addresses the two main forms of structural corruption. The Gaussian feature-neighborhood repair term introduces new propagation paths between feature-consistent nodes, which helps when useful edges are missing. The topology-aware residual term preserves the observed topology but adjusts its influence according to local feature and structural evidence, which helps reduce over-reliance on unreliable observed edges. As a result, TAGR produces a sparse repaired graph that combines semantic feature-neighborhood information with the original graph structure.

\subsection{Normalization and Message Passing}
\label{subsec:normalization_message_passing}

The repaired adjacency matrix $\widetilde{\mathbf{A}}$ is used as the message-passing graph for a downstream GNN. For a GCN backbone, self-loops are first added:
\[
    \widehat{\mathbf{A}}
    =
    \widetilde{\mathbf{A}}+\mathbf{I}_n,
\]
where $\mathbf{I}_n$ is the identity matrix. Let
\[
    \widehat{\mathbf{D}}_{ii}
    =
    \sum_{j=1}^{n}\widehat{A}_{ij}
\]
be the weighted degree matrix. The normalized repaired adjacency is
\[
    \overline{\mathbf{A}}
    =
    \widehat{\mathbf{D}}^{-\frac{1}{2}}
    \widehat{\mathbf{A}}
    \widehat{\mathbf{D}}^{-\frac{1}{2}}.
\]

A GCN layer operating on the repaired graph is then defined as
\[
    \mathbf{H}^{(\ell+1)}
    =
    \phi\left(
        \overline{\mathbf{A}}
        \mathbf{H}^{(\ell)}
        \mathbf{W}^{(\ell)}
    \right),
\]
where $\mathbf{H}^{(0)}=\mathbf{X}$, $\mathbf{W}^{(\ell)}$ is a trainable weight matrix, and $\phi(\cdot)$ is a nonlinear activation function. For other message-passing architectures, such as GAT or GraphSAGE, the original edge set is replaced by the support of $\widetilde{\mathbf{A}}$. TAGR therefore changes only the graph used for propagation; it does not change the downstream GNN architecture.

\subsection{Training Objective}
\label{subsec:training_objective}

Let $f_\theta(\mathbf{X},\widetilde{\mathbf{A}})$ denote a GNN with parameters $\theta$ operating on the repaired graph. For semi-supervised node classification, the model outputs class probabilities
\[
    \mathbf{Z}
    =
    \operatorname{softmax}
    \left(
        f_\theta(\mathbf{X},\widetilde{\mathbf{A}})
    \right),
\]
where $Z_{ic}$ is the predicted probability that node $v_i$ belongs to class $c$.

Let $\mathcal{V}_L\subset\mathcal{V}$ be the set of labeled nodes and let $\mathbf{Y}\in\{0,1\}^{n\times C}$ be the one-hot label matrix. The model is trained using the cross-entropy loss over labeled nodes:
\[
    \mathcal{L}_{\mathrm{cls}}
    =
    -
    \sum_{i\in\mathcal{V}_L}
    \sum_{c=1}^{C}
    Y_{ic}\log Z_{ic}.
\]
The optimization problem is
\[
    \theta^{*}
    =
    \arg\min_{\theta}
    \mathcal{L}_{\mathrm{cls}}
    \left(
        f_\theta(\mathbf{X},\widetilde{\mathbf{A}}),
        \mathbf{Y}
    \right).
\]

Since $\widetilde{\mathbf{A}}$ is constructed before GNN training, the learning objective remains the same as standard supervised GNN training. TAGR therefore avoids bilevel optimization and does not introduce additional trainable graph-generator parameters. It can be understood as a sparse repair layer placed before message passing: rather than replacing the GNN architecture, it improves the graph on which the architecture operates.

\subsection{Computational Complexity}
\label{subsec:complexity}

TAGR is designed to avoid dense graph structure learning. The Gaussian feature-neighborhood repair component selects $k$ candidate edges per node, producing $O(kn)$ directed candidates before symmetrization. The topology-aware residual is computed only on observed edges, requiring edge-level statistics over $|\mathcal{E}_A|$ edges. Therefore, the support of the repaired graph satisfies
\[
    |\operatorname{supp}(\widetilde{\mathbf{A}})|
    =
    O(|\mathcal{E}_A|+kn).
\]
When $k\ll n$, this is substantially smaller than dense graph learning over $O(n^2)$ possible node pairs.

Once the repaired graph is constructed, downstream training has the same form as ordinary message passing on a sparse weighted graph. Thus, TAGR improves the propagation structure without adding a graph-learning optimization loop or requiring dense adjacency parameters. This makes the method modular, efficient, and compatible with standard GNN backbones.

\section{Experiments}
\label{sec:experiments}

We evaluate TAGR on semi-supervised node classification under both clean and structurally corrupted graph settings. The goal is to examine whether sparse graph repair can improve the reliability of GNN message passing when the observed topology contains either spurious or missing edges. Specifically, the experiments assess: (i) robustness under edge-addition and edge-deletion perturbations; (ii) the contribution of Gaussian feature-neighborhood repair and topology-aware residual reweighting; (iii) compatibility with different GNN backbones; and (iv) comparison with representative graph repair and graph structure learning baselines.

\subsection{Experimental Setup}
\label{subsec:experimental_setup}

\paragraph{Datasets.}
We use four citation-network benchmarks: Cora, Citeseer, Cora-ML, and Pubmed. These datasets are widely used in semi-supervised node classification and differ in graph size, feature dimensionality, number of classes, and edge density. Table~\ref{tab:dataset_statistics} summarizes their statistics. Following the common transductive evaluation setting, each dataset is split into labeled training nodes, validation nodes, and test nodes. We use 20 labeled nodes per class for training, 500 nodes for validation, and 1,000 nodes for testing. The validation set is used for early stopping and hyperparameter selection, and final performance is reported on the test set.

\begin{table}[t]
\centering
\caption{The statistics of the datasets.}
\label{tab:dataset_statistics}
\begin{tabular}{lrrrrrrr}
\toprule
Dataset & Classes & Nodes & Edges & Features & Train & Validation & Test \\
\midrule
Cora & 7 & 2,708 & 5,278 & 1,433 & 140 & 500 & 1,000 \\
Citeseer & 6 & 3,312 & 4,536 & 3,703 & 120 & 500 & 1,000 \\
Cora-ML & 7 & 2,995 & 8,158 & 2,879 & 140 & 500 & 1,000 \\
Pubmed & 3 & 19,717 & 44,324 & 500 & 60 & 500 & 1,000 \\
\bottomrule
\end{tabular}
\end{table}

\paragraph{Baselines.}
We compare TAGR with three groups of methods. The first group consists of standard GNN backbones trained directly on the observed graph, including GCN, GAT, and GraphSAGE. These baselines measure the sensitivity of ordinary message passing to structural corruption. The second group contains graph repair variants built on the same backbones. Gaussian-only repair adds the Gaussian feature-neighborhood graph but does not use topology-aware residual reweighting, which isolates the contribution of feature-based repair. TAGR applies the full proposed repair mechanism. We also include RS-style \cite{dai2022towards} learned graph repair with matched GNN backbones, since learned denoising and densification are strong competitors under structural perturbations. The third group includes JNSGSL \cite{liu2026graph}, a recent graph structure learning method that jointly exploits node and structural feature representations. All methods are evaluated using the same data splits, node features, corrupted graphs, and model-selection protocol whenever computationally feasible.

\paragraph{Graph corruption protocol.}
We consider three graph conditions. The raw graph setting uses the original observed graph without artificial perturbation. The edge-addition setting simulates noisy topology by randomly inserting non-existing edges into the graph. In the main robustness table, we report Add 50\% and Add 90\%, where the number of added edges equals 50\% and 90\% of the original edge count. In the edge-deletion setting, we randomly remove observed edges to simulate missing or incomplete topology. The main table reports Del 25\% and Del 50\%. For robustness-curve analysis, the edge-addition ratio is varied from 0.1 to 0.9, and the edge-deletion ratio is varied from 0.05 to 0.50. These two corruption types test complementary failure modes: edge addition evaluates resistance to spurious message-passing paths, while edge deletion evaluates the ability to recover useful propagation under missing connectivity.

\paragraph{Implementation details.}
All experiments are conducted under a common benchmark pipeline. For each dataset, corruption setting, and random seed, all methods use the same node features, labels, train/validation/test split, and corrupted adjacency matrix. We select model checkpoints according to the best validation accuracy and report the corresponding test accuracy. TAGR hyperparameters are selected on the clean validation graph for each dataset and backbone and then fixed across all corruption settings. This prevents tuning directly on corrupted test conditions and evaluates whether the selected repair parameters generalize to structural perturbations. Unless otherwise stated, results are averaged over five random seeds, and we report mean test accuracy with standard deviation.

\begin{sidewaystable*}[htbp!]
\centering
\scriptsize
\definecolor{bestgreen}{HTML}{2E7D32}
\definecolor{secondblue}{HTML}{1565C0}
\definecolor{thirdorange}{HTML}{EF6C00}
\setlength{\tabcolsep}{3pt}
\caption{Main robustness comparison under the common protocol. Results are test accuracy (\%) averaged over five seeds, shown as mean $\pm$ standard deviation. Best, second-best, and third-best results in each row are shown in green, blue, and orange, respectively.}
\label{tab:main_joint_results}
\resizebox{\textwidth}{!}{%
\begin{tabular}{llcccc|cccc|cccc|c}
\toprule
Dataset & Graph & GCN & G-GCN & TAGR-GCN & RS-GCN & GAT & G-GAT & TAGR-GAT & RS-GAT & GS & G-GS & TAGR-GS & RS-GS & JNSGSL \\
\midrule
Cora & Raw Graph & \textcolor{thirdorange}{79.9 $\pm$ 0.7} & \textcolor{thirdorange}{79.9 $\pm$ 1.3} & \textcolor{bestgreen}{\textbf{80.5 $\pm$ 0.7}} & 79.3 $\pm$ 0.6 & 79.5 $\pm$ 1.4 & 76.8 $\pm$ 1.8 & 76.8 $\pm$ 1.8 & 67.8 $\pm$ 2.2 & 76.5 $\pm$ 0.8 & 76.6 $\pm$ 0.7 & 76.6 $\pm$ 0.6 & 72.9 $\pm$ 2.0 & \textcolor{secondblue}{80.5 $\pm$ 0.7} \\
 & Add 50\% & 72.6 $\pm$ 1.7 & \textcolor{thirdorange}{75.3 $\pm$ 1.4} & 75.1 $\pm$ 1.1 & \textcolor{bestgreen}{\textbf{76.5 $\pm$ 1.2}} & 70.7 $\pm$ 3.2 & \textcolor{secondblue}{75.6 $\pm$ 1.9} & \textcolor{secondblue}{75.6 $\pm$ 1.9} & 67.7 $\pm$ 2.3 & 71.2 $\pm$ 1.3 & 72.0 $\pm$ 1.5 & 72.5 $\pm$ 1.4 & 68.8 $\pm$ 1.8 & 74.1 $\pm$ 1.6 \\
 & Add 90\% & 67.9 $\pm$ 1.1 & 72.3 $\pm$ 1.0 & \textcolor{thirdorange}{72.3 $\pm$ 1.2} & \textcolor{bestgreen}{\textbf{74.8 $\pm$ 1.5}} & 66.6 $\pm$ 3.5 & \textcolor{secondblue}{73.6 $\pm$ 1.8} & \textcolor{secondblue}{73.6 $\pm$ 1.8} & 67.6 $\pm$ 2.0 & 67.8 $\pm$ 1.5 & 68.4 $\pm$ 1.1 & 69.2 $\pm$ 1.2 & 66.3 $\pm$ 1.4 & 70.1 $\pm$ 2.1 \\
 & Del 25\% & 78.0 $\pm$ 0.7 & \textcolor{secondblue}{79.4 $\pm$ 0.9} & \textcolor{bestgreen}{\textbf{79.6 $\pm$ 1.0}} & 77.7 $\pm$ 0.5 & 76.2 $\pm$ 1.7 & 74.5 $\pm$ 1.3 & 74.5 $\pm$ 1.3 & 67.1 $\pm$ 2.2 & 74.4 $\pm$ 1.6 & 75.1 $\pm$ 1.0 & 74.9 $\pm$ 0.9 & 69.8 $\pm$ 2.5 & \textcolor{thirdorange}{78.7 $\pm$ 1.5} \\
 & Del 50\% & 72.7 $\pm$ 1.6 & \textcolor{secondblue}{74.9 $\pm$ 1.4} & \textcolor{bestgreen}{\textbf{75.2 $\pm$ 1.5}} & 74.4 $\pm$ 1.5 & 70.1 $\pm$ 2.5 & 71.5 $\pm$ 1.0 & 71.5 $\pm$ 1.0 & 66.5 $\pm$ 1.6 & 69.6 $\pm$ 1.7 & 71.2 $\pm$ 0.9 & 71.1 $\pm$ 1.1 & 66.3 $\pm$ 2.0 & \textcolor{thirdorange}{74.4 $\pm$ 2.1} \\
\midrule
Cora-ML & Raw Graph & 82.7 $\pm$ 2.1 & \textcolor{secondblue}{83.5 $\pm$ 2.0} & \textcolor{bestgreen}{\textbf{83.6 $\pm$ 2.0}} & 80.5 $\pm$ 2.3 & 79.9 $\pm$ 2.6 & 77.8 $\pm$ 2.3 & 77.8 $\pm$ 2.3 & 71.8 $\pm$ 2.6 & 80.0 $\pm$ 1.7 & 79.7 $\pm$ 1.1 & 79.3 $\pm$ 1.3 & 73.5 $\pm$ 1.8 & \textcolor{thirdorange}{82.8 $\pm$ 1.8} \\
 & Add 50\% & 74.0 $\pm$ 2.2 & 75.5 $\pm$ 2.5 & 76.1 $\pm$ 1.7 & \textcolor{bestgreen}{\textbf{78.1 $\pm$ 1.3}} & 69.9 $\pm$ 2.7 & 73.5 $\pm$ 1.9 & 73.5 $\pm$ 1.9 & 72.3 $\pm$ 2.6 & 75.8 $\pm$ 1.5 & \textcolor{thirdorange}{76.7 $\pm$ 1.3} & \textcolor{secondblue}{77.3 $\pm$ 1.9} & 72.1 $\pm$ 2.1 & 76.6 $\pm$ 1.4 \\
 & Add 90\% & 69.7 $\pm$ 3.2 & 72.0 $\pm$ 3.0 & 72.3 $\pm$ 2.8 & \textcolor{bestgreen}{\textbf{77.4 $\pm$ 1.4}} & 66.8 $\pm$ 3.5 & 73.1 $\pm$ 2.5 & 73.1 $\pm$ 2.5 & 71.8 $\pm$ 2.1 & 73.2 $\pm$ 1.3 & \textcolor{thirdorange}{74.2 $\pm$ 1.4} & \textcolor{secondblue}{75.3 $\pm$ 1.4} & 71.9 $\pm$ 2.1 & 73.6 $\pm$ 1.9 \\
 & Del 25\% & 80.3 $\pm$ 1.4 & \textcolor{thirdorange}{80.6 $\pm$ 2.0} & \textcolor{bestgreen}{\textbf{81.3 $\pm$ 2.5}} & 78.1 $\pm$ 1.9 & 76.8 $\pm$ 2.4 & 75.8 $\pm$ 2.4 & 75.8 $\pm$ 2.4 & 71.5 $\pm$ 2.6 & 77.3 $\pm$ 2.1 & 78.0 $\pm$ 1.2 & 77.6 $\pm$ 1.2 & 72.6 $\pm$ 1.9 & \textcolor{secondblue}{81.1 $\pm$ 1.7} \\
 & Del 50\% & 76.1 $\pm$ 1.9 & \textcolor{thirdorange}{78.1 $\pm$ 1.9} & \textcolor{secondblue}{78.4 $\pm$ 1.6} & 75.2 $\pm$ 1.3 & 73.1 $\pm$ 3.0 & 73.2 $\pm$ 2.3 & 73.2 $\pm$ 2.3 & 70.5 $\pm$ 2.2 & 75.1 $\pm$ 2.4 & 76.2 $\pm$ 1.9 & 76.5 $\pm$ 0.9 & 71.6 $\pm$ 1.8 & \textcolor{bestgreen}{\textbf{78.5 $\pm$ 1.9}} \\
\midrule
Citeseer & Raw Graph & 65.9 $\pm$ 1.8 & \textcolor{bestgreen}{\textbf{68.0 $\pm$ 1.4}} & \textcolor{secondblue}{68.0 $\pm$ 1.2} & \textcolor{thirdorange}{67.8 $\pm$ 1.0} & 65.8 $\pm$ 2.4 & 67.4 $\pm$ 2.0 & 67.4 $\pm$ 2.0 & 64.5 $\pm$ 2.2 & 65.1 $\pm$ 2.1 & 65.3 $\pm$ 1.4 & 65.3 $\pm$ 1.5 & 63.3 $\pm$ 0.5 & 66.9 $\pm$ 2.2 \\
 & Add 50\% & 60.7 $\pm$ 1.8 & 63.3 $\pm$ 1.7 & 63.8 $\pm$ 1.7 & \textcolor{bestgreen}{\textbf{66.8 $\pm$ 1.8}} & 57.8 $\pm$ 1.3 & \textcolor{secondblue}{65.1 $\pm$ 2.7} & \textcolor{secondblue}{65.1 $\pm$ 2.7} & \textcolor{thirdorange}{65.0 $\pm$ 1.9} & 60.7 $\pm$ 2.0 & 61.6 $\pm$ 1.4 & 61.7 $\pm$ 1.6 & 63.9 $\pm$ 1.0 & 61.0 $\pm$ 1.8 \\
 & Add 90\% & 56.5 $\pm$ 0.9 & 60.5 $\pm$ 1.1 & 61.3 $\pm$ 1.0 & \textcolor{bestgreen}{\textbf{66.4 $\pm$ 1.1}} & 54.0 $\pm$ 1.0 & \textcolor{secondblue}{64.7 $\pm$ 2.2} & \textcolor{secondblue}{64.7 $\pm$ 2.2} & \textcolor{thirdorange}{63.6 $\pm$ 3.2} & 58.7 $\pm$ 0.7 & 59.4 $\pm$ 0.8 & 59.6 $\pm$ 0.9 & 63.2 $\pm$ 0.8 & 57.3 $\pm$ 1.6 \\
 & Del 25\% & 65.3 $\pm$ 2.4 & \textcolor{bestgreen}{\textbf{67.3 $\pm$ 2.2}} & \textcolor{secondblue}{67.2 $\pm$ 2.1} & 66.8 $\pm$ 1.2 & 62.8 $\pm$ 2.6 & 66.5 $\pm$ 2.2 & 66.5 $\pm$ 2.2 & 65.0 $\pm$ 1.8 & 62.5 $\pm$ 1.8 & 64.8 $\pm$ 2.1 & 64.1 $\pm$ 2.3 & 63.1 $\pm$ 0.6 & \textcolor{thirdorange}{67.1 $\pm$ 2.6} \\
 & Del 50\% & 61.9 $\pm$ 1.4 & \textcolor{secondblue}{66.4 $\pm$ 1.4} & \textcolor{bestgreen}{\textbf{66.4 $\pm$ 1.4}} & 64.8 $\pm$ 1.3 & 61.2 $\pm$ 1.5 & \textcolor{thirdorange}{65.4 $\pm$ 2.0} & \textcolor{thirdorange}{65.4 $\pm$ 2.0} & 64.4 $\pm$ 1.8 & 60.6 $\pm$ 1.5 & 61.5 $\pm$ 1.1 & 61.9 $\pm$ 0.7 & 63.5 $\pm$ 0.9 & 64.9 $\pm$ 1.4 \\
\midrule
Pubmed & Raw Graph & \textcolor{secondblue}{76.5 $\pm$ 0.6} & 76.5 $\pm$ 1.3 & \textcolor{thirdorange}{76.5 $\pm$ 1.2} & 67.2 $\pm$ 2.0 & 71.7 $\pm$ 4.0 & 72.2 $\pm$ 2.8 & 72.2 $\pm$ 2.8 & 67.6 $\pm$ 2.4 & 72.9 $\pm$ 1.8 & 73.8 $\pm$ 1.6 & 73.6 $\pm$ 1.2 & 70.1 $\pm$ 2.3 & \textcolor{bestgreen}{\textbf{76.7 $\pm$ 1.6}} \\
 & Add 50\% & 70.0 $\pm$ 1.8 & \textcolor{secondblue}{72.8 $\pm$ 1.9} & \textcolor{bestgreen}{\textbf{72.9 $\pm$ 1.9}} & 67.1 $\pm$ 1.4 & 65.9 $\pm$ 3.7 & 68.4 $\pm$ 3.6 & 68.4 $\pm$ 3.6 & 66.7 $\pm$ 2.6 & 70.0 $\pm$ 2.3 & \textcolor{thirdorange}{71.5 $\pm$ 1.1} & 71.3 $\pm$ 1.2 & 69.9 $\pm$ 2.4 & 69.7 $\pm$ 1.8 \\
 & Add 90\% & 68.1 $\pm$ 1.5 & \textcolor{secondblue}{70.0 $\pm$ 1.7} & \textcolor{bestgreen}{\textbf{70.2 $\pm$ 1.4}} & 67.4 $\pm$ 1.5 & 63.1 $\pm$ 2.4 & 68.3 $\pm$ 3.8 & 68.3 $\pm$ 3.8 & 67.1 $\pm$ 2.7 & 68.8 $\pm$ 0.9 & \textcolor{thirdorange}{70.0 $\pm$ 1.2} & \textcolor{bestgreen}{\textbf{70.2 $\pm$ 1.3}} & 69.6 $\pm$ 2.3 & 67.5 $\pm$ 2.1 \\
 & Del 25\% & 74.4 $\pm$ 0.9 & \textcolor{thirdorange}{75.0 $\pm$ 1.1} & \textcolor{secondblue}{75.2 $\pm$ 1.0} & 66.3 $\pm$ 1.6 & 71.0 $\pm$ 2.9 & 70.2 $\pm$ 2.7 & 70.2 $\pm$ 2.7 & 66.8 $\pm$ 2.3 & 72.4 $\pm$ 1.2 & 73.2 $\pm$ 1.6 & 73.0 $\pm$ 1.5 & 69.5 $\pm$ 2.3 & \textcolor{bestgreen}{\textbf{75.9 $\pm$ 2.5}} \\
 & Del 50\% & \textcolor{thirdorange}{73.1 $\pm$ 0.6} & \textcolor{secondblue}{73.2 $\pm$ 1.4} & \textcolor{bestgreen}{\textbf{73.5 $\pm$ 0.8}} & 65.5 $\pm$ 2.4 & 69.3 $\pm$ 1.6 & 69.1 $\pm$ 3.6 & 69.1 $\pm$ 3.6 & 66.3 $\pm$ 2.2 & 70.9 $\pm$ 1.0 & 72.0 $\pm$ 1.2 & 72.0 $\pm$ 1.3 & 69.6 $\pm$ 2.5 & 73.0 $\pm$ 1.9 \\
\bottomrule
\end{tabular}%
}
\end{sidewaystable*}
\subsection{Main Robustness Comparison}
\label{subsec:main_results}

Table~\ref{tab:main_joint_results} reports the main robustness comparison under the common evaluation protocol. Overall, the results show that repairing the message-passing graph improves robustness across a wide range of structural conditions. The improvement is especially clear when the observed graph is corrupted by either edge addition or edge deletion, confirming that structural repair is useful for both noisy and incomplete topology.

The strongest and most consistent gains appear in the GCN family. This is expected because GCN directly propagates features through the normalized adjacency matrix, making it highly sensitive to the quality of the graph used for message passing. When random edges are added, irrelevant neighbors can inject noisy information into node representations; when edges are removed, useful propagation paths disappear. TAGR-GCN mitigates both effects by augmenting the graph with feature-consistent Gaussian neighborhoods and reweighting the observed topology using local feature-structural evidence. For example, on Cora, vanilla GCN drops from $79.9\%$ on the raw graph to $67.9\%$ under Add 90\%, whereas TAGR-GCN obtains $72.3\%$ under the same perturbation. On Citeseer, GCN reaches only $56.5\%$ under Add 90\%, while TAGR-GCN improves the result to $61.3\%$. Under edge deletion, TAGR-GCN also remains consistently stronger than GCN; for instance, on Pubmed under Del 50\%, TAGR-GCN achieves $73.5\%$ compared with $73.1\%$ for GCN, and on Citeseer under Del 50\%, it improves from $61.9\%$ to $66.4\%$. These results indicate that TAGR reduces the dependence of GCN on the reliability of the originally observed adjacency matrix.

The comparison between GCN and G-GCN (Gaussian only) shows that feature-neighborhood repair is already a strong robustness mechanism. In many corrupted settings, G-GCN substantially improves over vanilla GCN. This confirms that node features contain useful relational information that is not fully captured by the observed graph. For example, on Cora under Add 90\%, G-GCN improves GCN from $67.9\%$ to $72.3\%$, and on Citeseer under Del 50\%, it improves from $61.9\%$ to $66.4\%$. These gains support the central motivation of TAGR: when graph topology is noisy or incomplete, sparse feature-consistent auxiliary edges can provide a more reliable propagation structure.

The full TAGR model usually matches or improves over Gaussian-only repair, with the clearest benefits appearing under deletion noise and on larger or more structurally incomplete graphs. On Cora-ML under Del 25\%, TAGR-GCN improves over G-GCN from $80.6\%$ to $81.3\%$, and under Del 50\% from $78.1\%$ to $78.4\%$. On Pubmed, TAGR-GCN also improves over G-GCN under both Add 90\% and Del 50\%. Although the numerical gains over Gaussian-only repair are sometimes modest, they are meaningful because the topology-aware residual is not intended to replace feature repair; rather, it stabilizes the repaired graph by preserving and adjusting useful observed edges. This behavior is consistent with the design of TAGR, where the Gaussian feature-neighborhood graph restores missing semantic connections, while the topology-aware residual refines the original topology.

The comparison with RS-style learned graph repair reveals an important distinction between learned and deterministic repair strategies. RS-GCN is particularly strong under severe edge-addition noise on smaller citation graphs. It achieves the best result on Cora, Cora-ML, and Citeseer under Add 90\%, suggesting that learned graph repair can be highly effective when the main corruption is the insertion of spurious edges. However, TAGR-GCN remains competitive while using a simpler sparse repair mechanism that does not require a learned graph generator or dense pairwise graph optimization. This distinction is practically important: TAGR sacrifices some adaptivity in high-noise addition settings, but gains simplicity, modularity, and stable behavior across different corruption regimes.

JNSGSL also performs strongly in several clean and moderately corrupted settings, which is expected because it explicitly learns graph structure from both node and structural representations. However, its performance under severe edge addition is less stable on some datasets. For example, under Add 90\%, JNSGSL obtains $70.1\%$ on Cora and $57.3\%$ on Citeseer, while TAGR-GCN reaches $72.3\%$ and $61.3\%$, respectively. This suggests that more expressive graph structure learning does not automatically lead to stronger robustness under every perturbation type. In contrast, TAGR relies on a constrained repair mechanism that adds sparse feature-consistent edges and preserves the observed graph through bounded residual reweighting, which can provide more stable behavior when the corruption regime is unknown.

The effect of repair also depends on the downstream backbone. For GCN, the benefit is direct because the repaired weighted adjacency is exactly the matrix used in propagation. This explains why TAGR-GCN and G-GCN show the clearest improvements. For GAT, Gaussian repair and TAGR often produce similar results. This is reasonable because GAT already learns attention weights over neighbors, which can partially absorb edge-weight differences introduced by the residual reweighting. Nevertheless, repairing the candidate edge set remains important: on Citeseer under Add 90\%, GAT obtains $54.0\%$, while G-GAT and TAGR-GAT both reach $64.7\%$. Thus, even when attention reduces sensitivity to explicit edge weights, adding better feature-neighborhood candidates can substantially improve robustness. For GraphSAGE, the gains are more moderate but still visible in several noisy settings. On Cora-ML under Add 90\%, GraphSAGE obtains $73.2\%$, whereas TAGR-GraphSAGE reaches $75.3\%$. This indicates that the repaired graph remains useful across architectures, although the magnitude of improvement depends on how each backbone uses neighborhood information.

Taken together, the results in Table~\ref{tab:main_joint_results} support three conclusions. First, graph repair improves robustness over standard message passing, particularly for GCN. Second, Gaussian feature-neighborhood repair is the dominant source of improvement, showing that feature geometry provides reliable auxiliary structure under graph corruption. Third, topology-aware residual reweighting provides additional stabilization by incorporating local structural evidence from the observed graph. TAGR therefore offers a lightweight alternative to dense or fully learned graph structure learning: it is not always the best method in every single high-noise setting, but it provides strong and stable robustness across datasets, perturbation types, and GNN backbones.

\subsection{Robustness Curves}
\label{subsec:robustness_curves}

\begin{figure*}[!ht]
\begin{center}

    \subfigure[Cora.]{%
        \includegraphics[scale=.56]{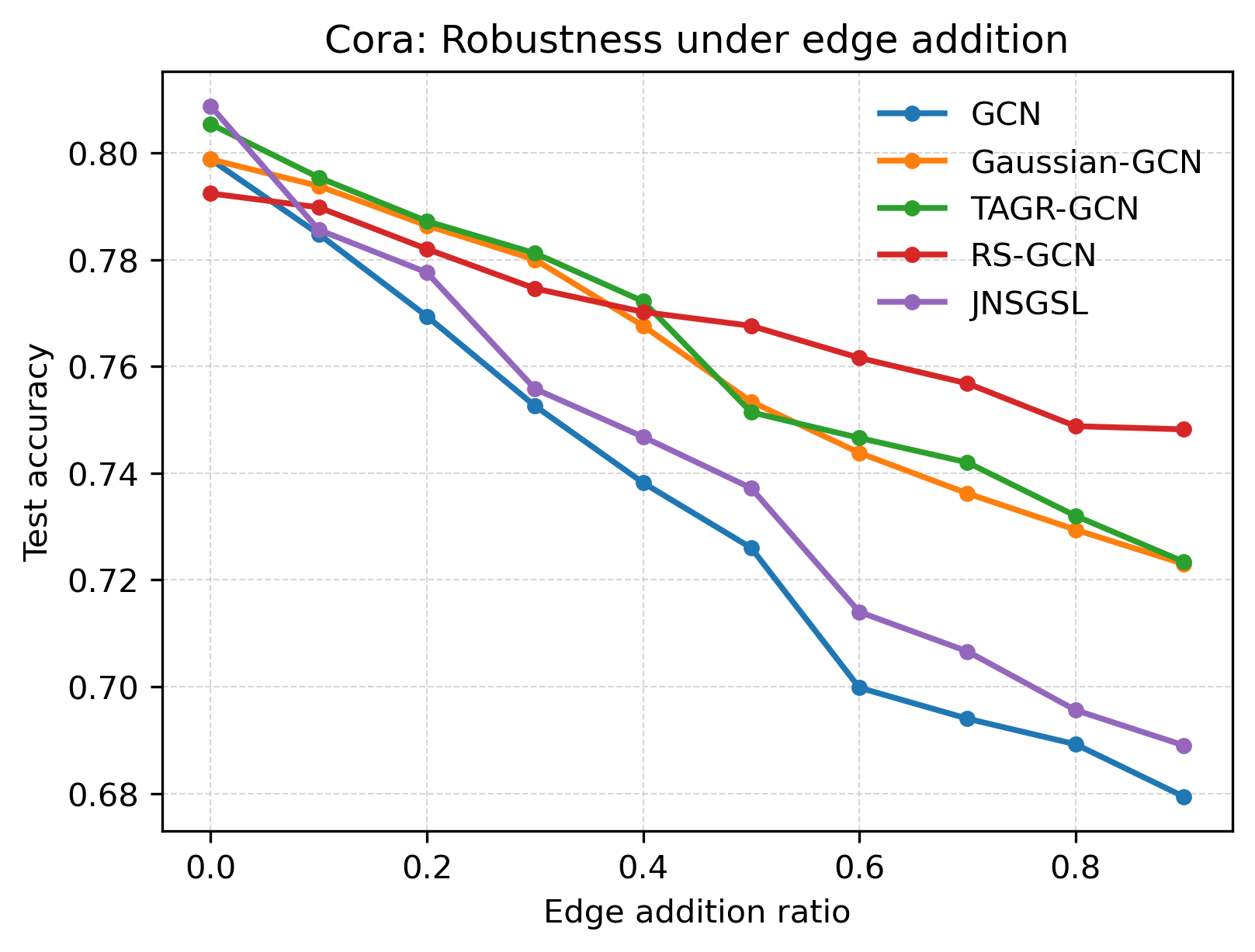}
        \label{fig:add_cora}}
    \quad
    \subfigure[Citeseer.]{%
        \includegraphics[scale=.56]{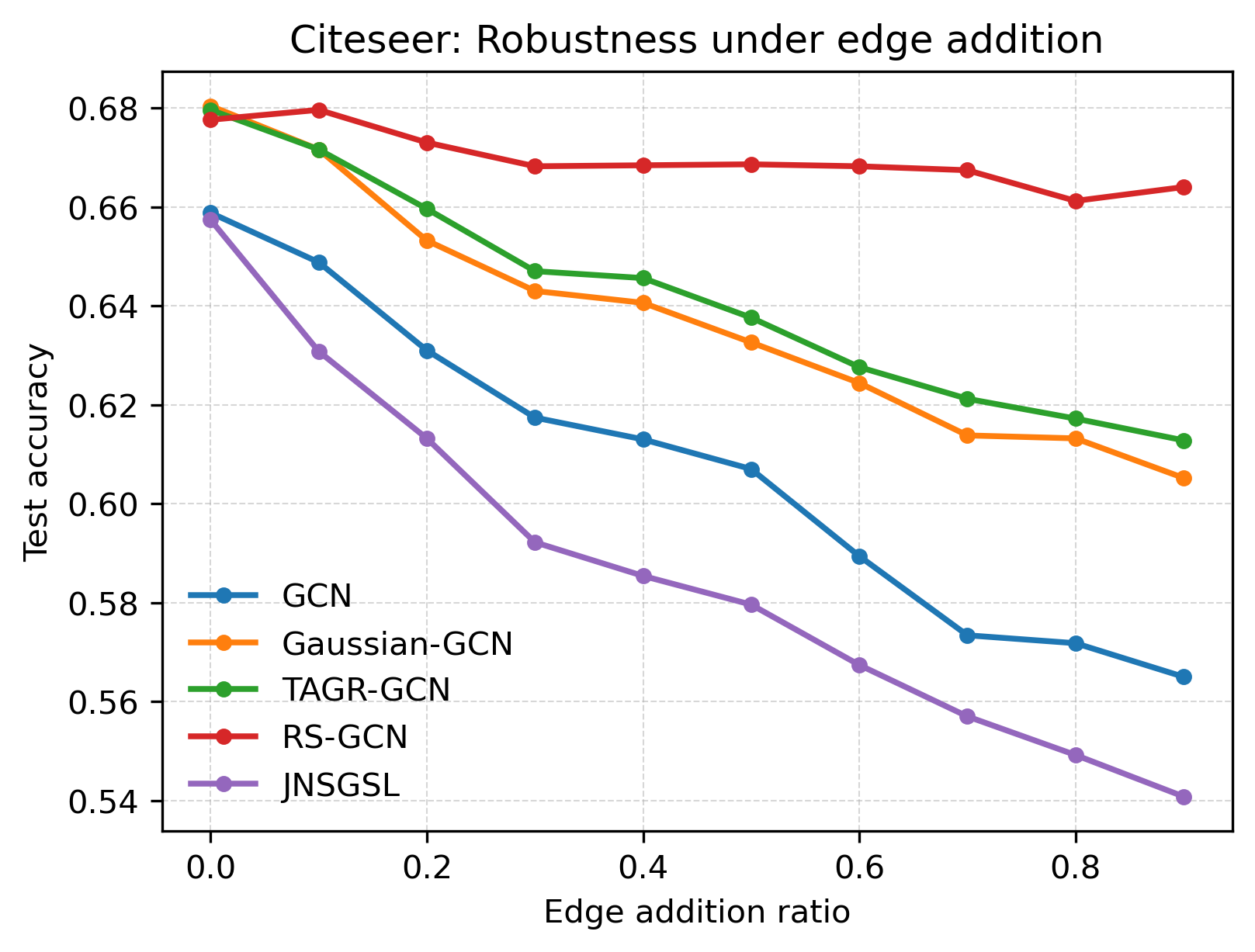}
        \label{fig:add_citeseer}}

    \vspace{0.5em}

    \subfigure[Cora-ML.]{%
        \includegraphics[scale=.56]{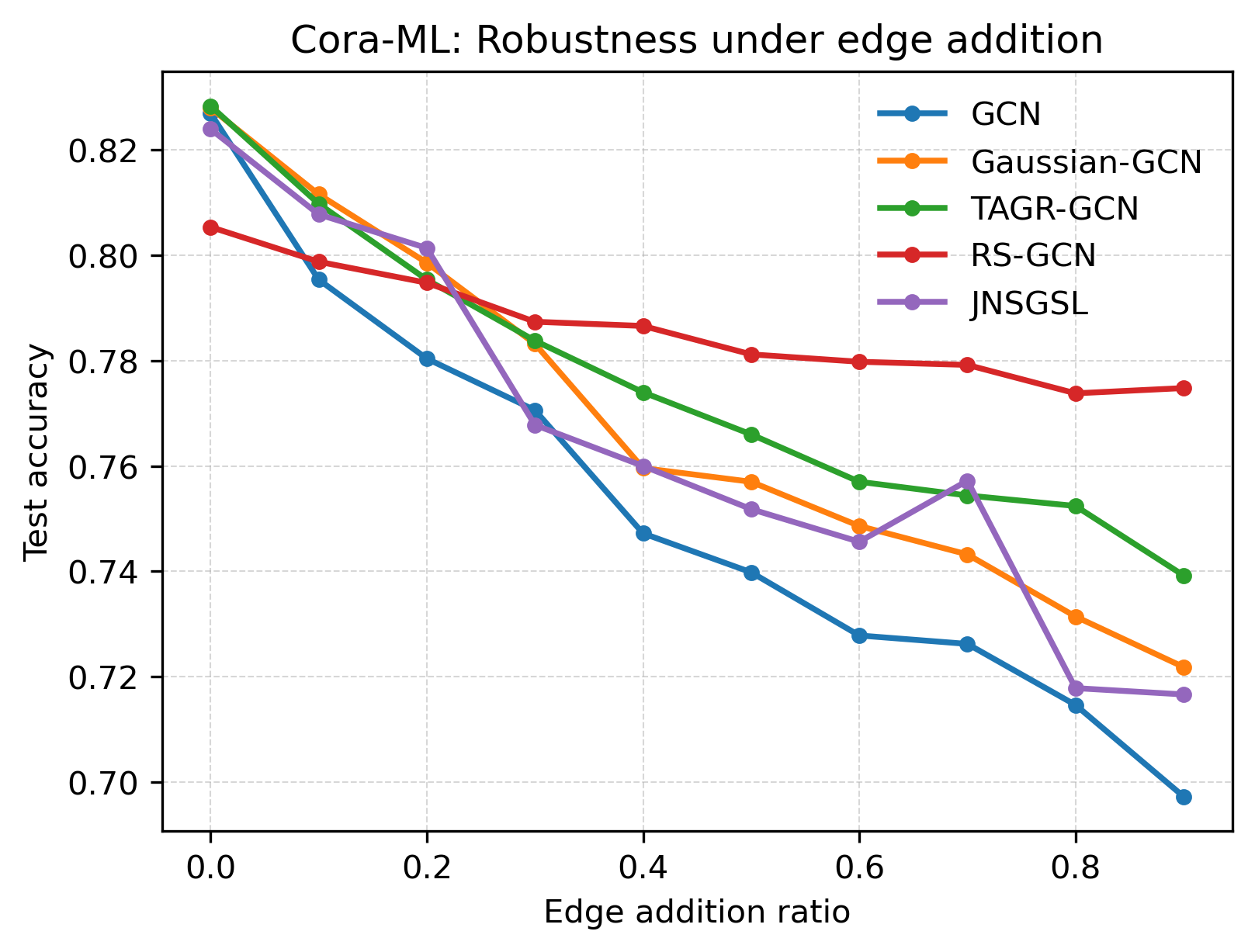}
        \label{fig:add_cora_ml}}
    \quad
    \subfigure[Pubmed.]{%
        \includegraphics[scale=.56]{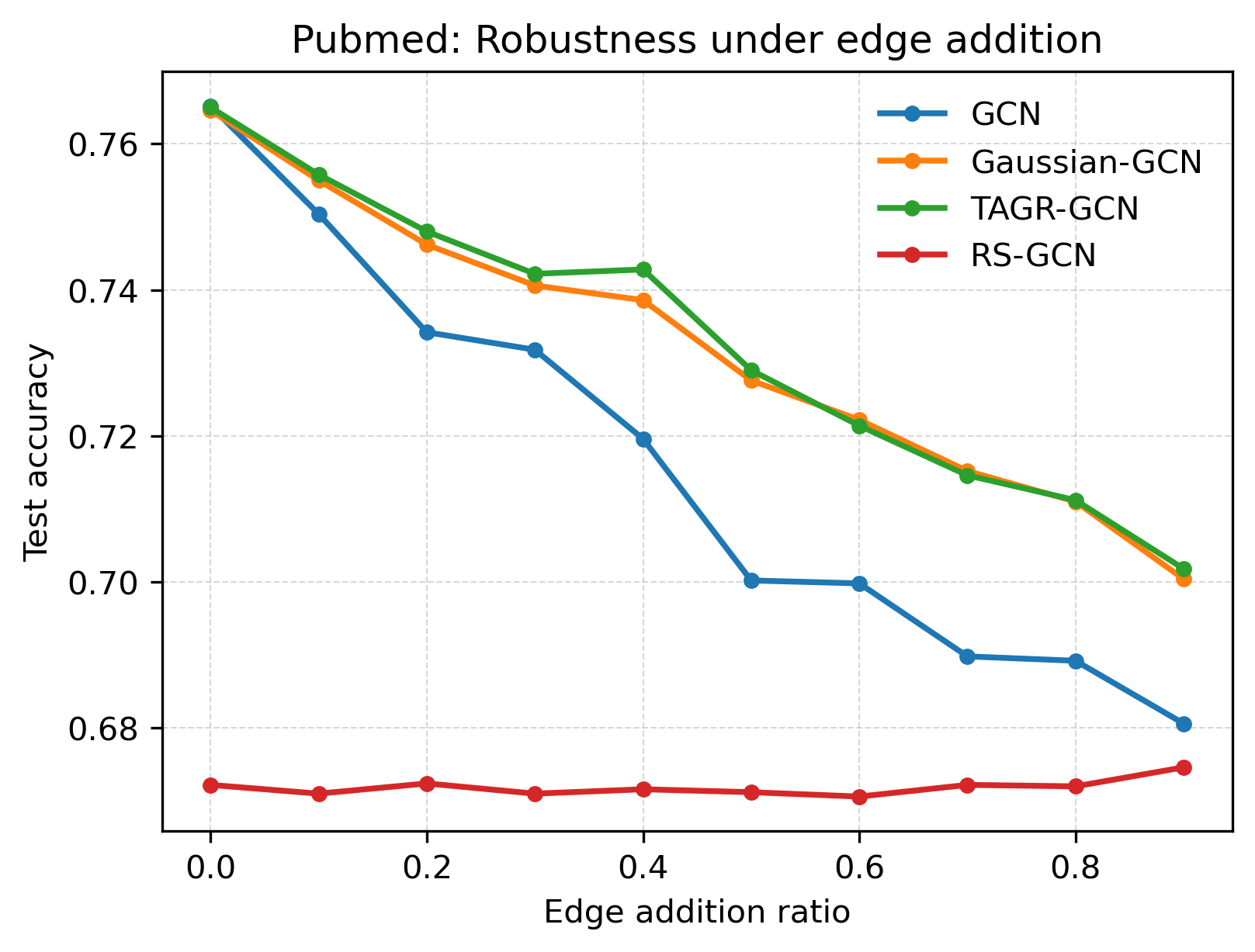}
        \label{fig:add_pubmed}}

\end{center}

\caption{Robustness under edge-addition noise with a fixed GCN backbone. The edge-addition ratio varies from $0.1$ to $0.9$, and each point reports the mean test accuracy over five random seeds. Vanilla GCN degrades rapidly as noisy edges are added, showing its sensitivity to spurious message-passing paths. Gaussian-GCN and TAGR-GCN consistently improve over GCN across all datasets by introducing feature-consistent repaired neighborhoods. RS-GCN is particularly strong on Cora, Cora-ML, and Citeseer under severe edge addition, while TAGR-GCN remains competitive and is especially effective on Pubmed.}
\label{fig:edge_addition_curves}

\end{figure*}

The main table reports representative corruption levels, but robustness should also be evaluated across a continuous range of perturbation strengths. We therefore fix the downstream backbone to GCN and compare GCN, Gaussian-GCN, TAGR-GCN, RS-GCN, and JNSGSL over full edge-addition and edge-deletion curves. For edge addition, the perturbation ratio varies from 0.1 to 0.9. For edge deletion, the deletion ratio varies from 0.05 to 0.50. This setting isolates the effect of graph repair under the same message-passing architecture.

Figure~\ref{fig:edge_addition_curves} shows the results under edge-addition noise. As the number of randomly inserted edges increases, vanilla GCN degrades substantially on all datasets. This confirms that spurious edges can create unreliable message-passing paths and contaminate node representations. Gaussian-GCN and TAGR-GCN consistently reduce this degradation, showing that feature-neighborhood repair provides an alternative propagation structure when the observed graph becomes increasingly noisy. The effect is especially clear on Pubmed, where both Gaussian-GCN and TAGR-GCN remain above vanilla GCN across almost the entire perturbation range.

The comparison with RS-GCN reveals that learned graph repair is particularly effective under severe edge-addition noise. On Cora, Citeseer, and Cora-ML, RS-GCN remains highly competitive and often achieves the best accuracy at high addition ratios. This suggests that a learned repair mechanism can be advantageous when the dominant corruption is the insertion of spurious edges. However, TAGR-GCN remains close to the strongest methods while using a deterministic sparse repair operator rather than a learned graph generator. This distinction is important because TAGR avoids additional graph-learning optimization and is directly compatible with standard GNN training.

\begin{figure*}[!ht]
\begin{center}

    \subfigure[Cora.]{%
        \includegraphics[scale=.56]{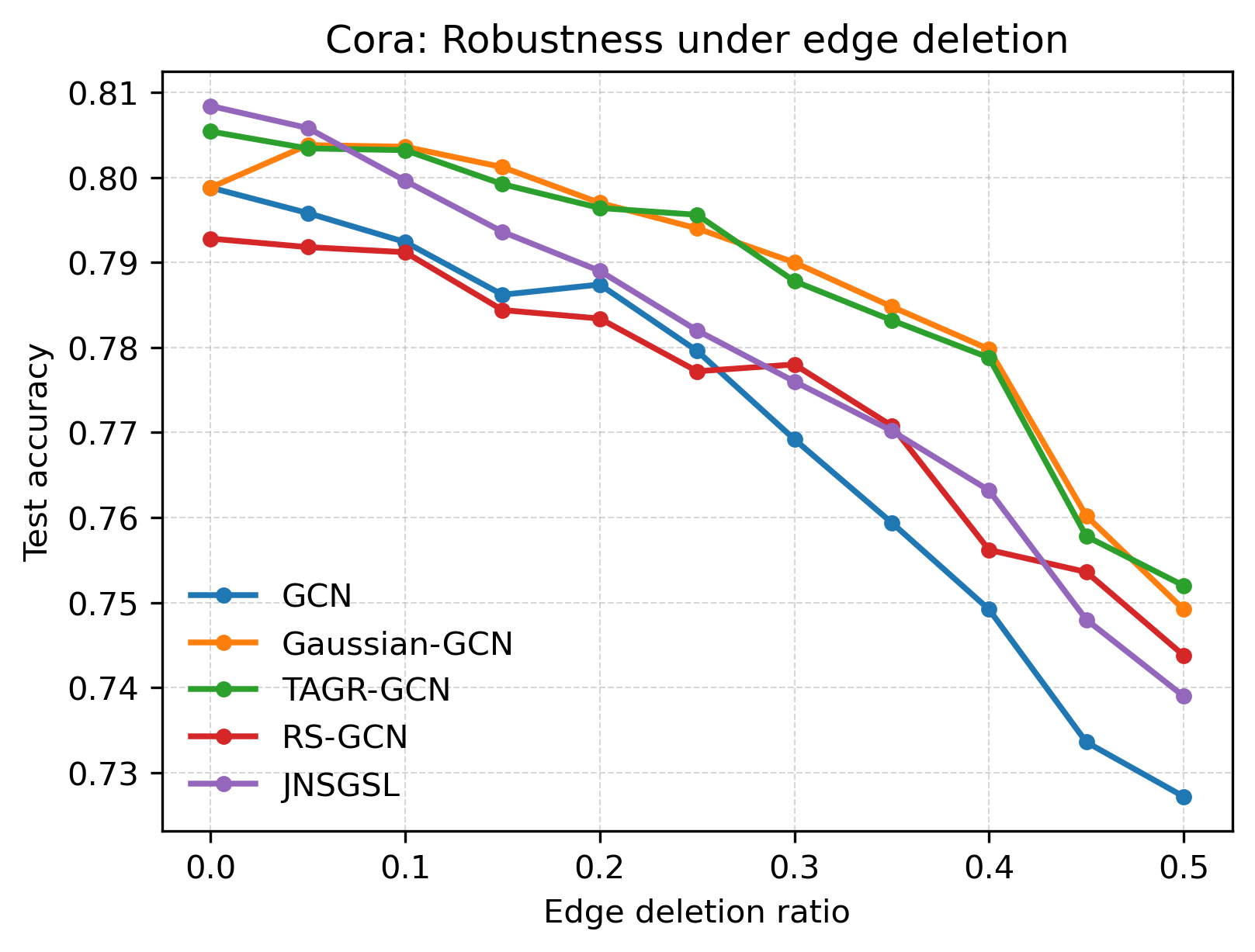}
        \label{fig:del_cora}}
    \quad
    \subfigure[Citeseer.]{%
        \includegraphics[scale=.56]{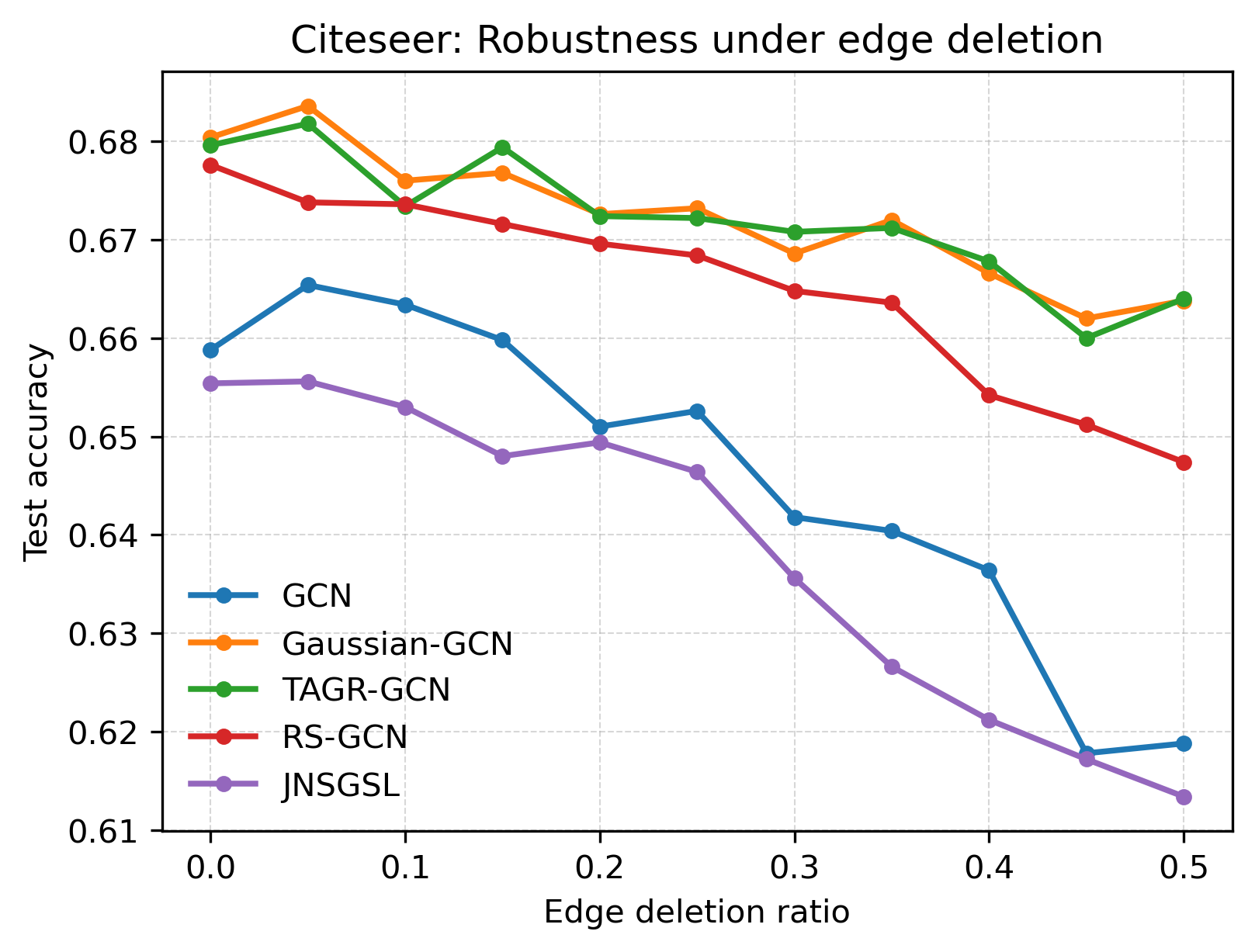}
        \label{fig:del_citeseer}}

    \vspace{0.5em}

    \subfigure[Cora-ML.]{%
        \includegraphics[scale=.56]{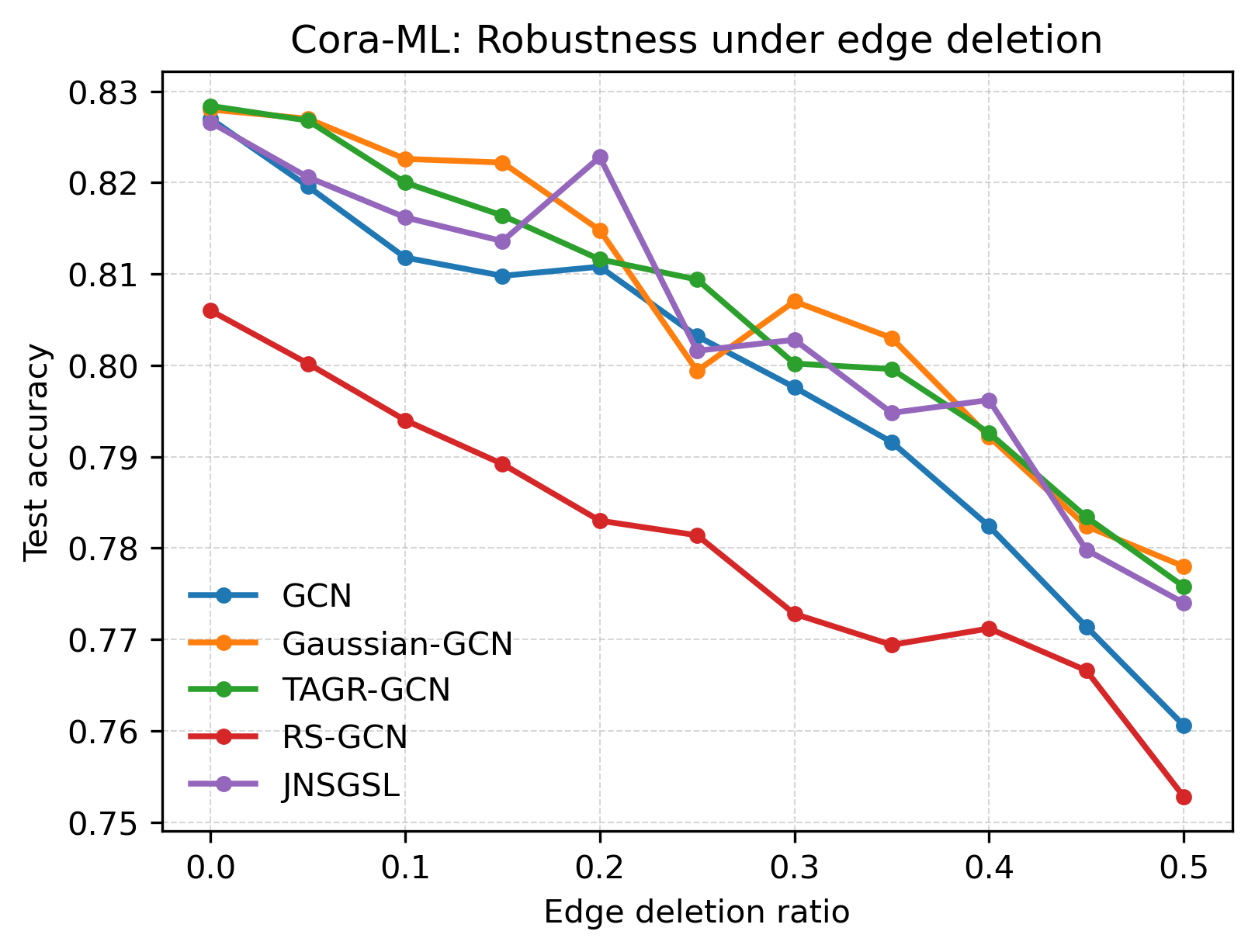}
        \label{fig:del_cora_ml}}
    \quad
    \subfigure[Pubmed.]{%
        \includegraphics[scale=.56]{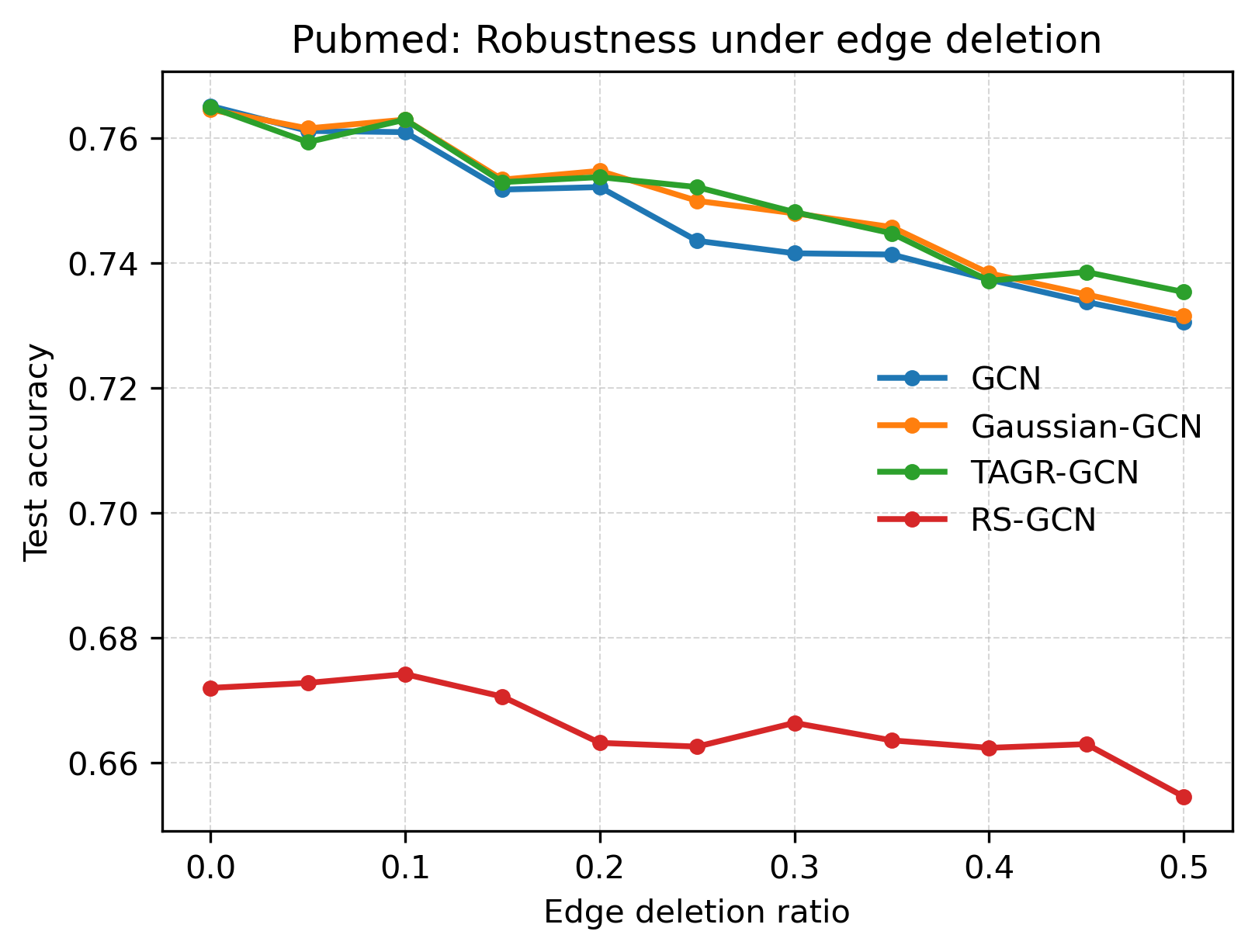}
        \label{fig:del_pubmed}}

\end{center}

\caption{Robustness under edge-deletion noise with a fixed GCN backbone. The edge-deletion ratio varies from $0.05$ to $0.50$, and each point reports the mean test accuracy over five random seeds. As edges are removed, vanilla GCN loses useful propagation paths and its performance decreases. Gaussian-GCN and TAGR-GCN consistently remain above GCN, indicating that feature-based graph repair can recover informative neighborhoods when the observed graph is incomplete. TAGR-GCN is particularly stable under high deletion ratios and remains among the strongest methods across the four datasets.}
\label{fig:edge_deletion_curves}

\end{figure*}

Figure~\ref{fig:edge_deletion_curves} evaluates robustness when observed edges are removed. This setting tests a different failure mode: the graph becomes incomplete, and useful propagation paths disappear. In this case, Gaussian-GCN and TAGR-GCN consistently outperform vanilla GCN across datasets. The improvement supports the main intuition behind Gaussian feature-neighborhood repair: when topology is missing, feature-consistent auxiliary edges can recover useful neighborhoods for message passing. TAGR-GCN is particularly stable under moderate and high deletion ratios, indicating that combining feature-neighborhood repair with topology-aware residual reweighting helps preserve robust propagation even as the observed graph becomes sparse.

JNSGSL is competitive in several clean and mildly corrupted settings, but its robustness varies more strongly across perturbation types and datasets. Under stronger edge addition, it degrades more rapidly than TAGR-GCN or RS-GCN on Cora and Citeseer. Under edge deletion, it remains competitive on Cora and Cora-ML but is less stable on Citeseer. We do not report JNSGSL on Pubmed in the robustness curves because its graph structure learning procedure was computationally expensive under our repeated perturbation protocol. This highlights a practical difference between dense or learned graph structure methods and lightweight sparse repair methods TAGR.

% \begin{figure*}[htbp!]
%     \centering
%     \begin{subfigure}[t]{0.48\textwidth}
%         \centering
%         \includegraphics[width=\linewidth]{gcn_fixed_del_cora.png}
%         \caption{Cora}
%         \label{fig:del_cora}
%     \end{subfigure}
%     \hfill
%     \begin{subfigure}[t]{0.48\textwidth}
%         \centering
%         \includegraphics[width=\linewidth]{gcn_fixed_del_citeseer.png}
%         \caption{Citeseer}
%         \label{fig:del_citeseer}
%     \end{subfigure}

%     \vspace{0.6em}

%     \begin{subfigure}[t]{0.48\textwidth}
%         \centering
%         \includegraphics[width=\linewidth]{gcn_fixed_del_cora_ml.png}
%         \caption{Cora-ML}
%         \label{fig:del_cora_ml}
%     \end{subfigure}
%     \hfill
%     \begin{subfigure}[t]{0.48\textwidth}
%         \centering
%         \includegraphics[width=\linewidth]{gcn_fixed_del_pubmed.png}
%         \caption{Pubmed}
%         \label{fig:del_pubmed}
%     \end{subfigure}

%     \caption{Robustness under edge deletion noise with a fixed GCN backbone. The edge deletion ratio varies from $0.05$ to $0.50$, and each point reports the mean test accuracy over five random seeds. As edges are removed, vanilla GCN loses useful propagation paths and its performance decreases. Gaussian-GCN and TAGR-GCN consistently remain above GCN, indicating that feature-based graph repair can recover informative neighborhoods when the observed graph is incomplete. TAGR-GCN is particularly stable under high deletion ratios and remains among the strongest methods across the four datasets.}
%     \label{fig:edge_deletion_curves}
% \end{figure*}

\begin{figure*}[!ht]
\begin{center}

    \subfigure[Edge addition.]{%
        \includegraphics[scale=.50]{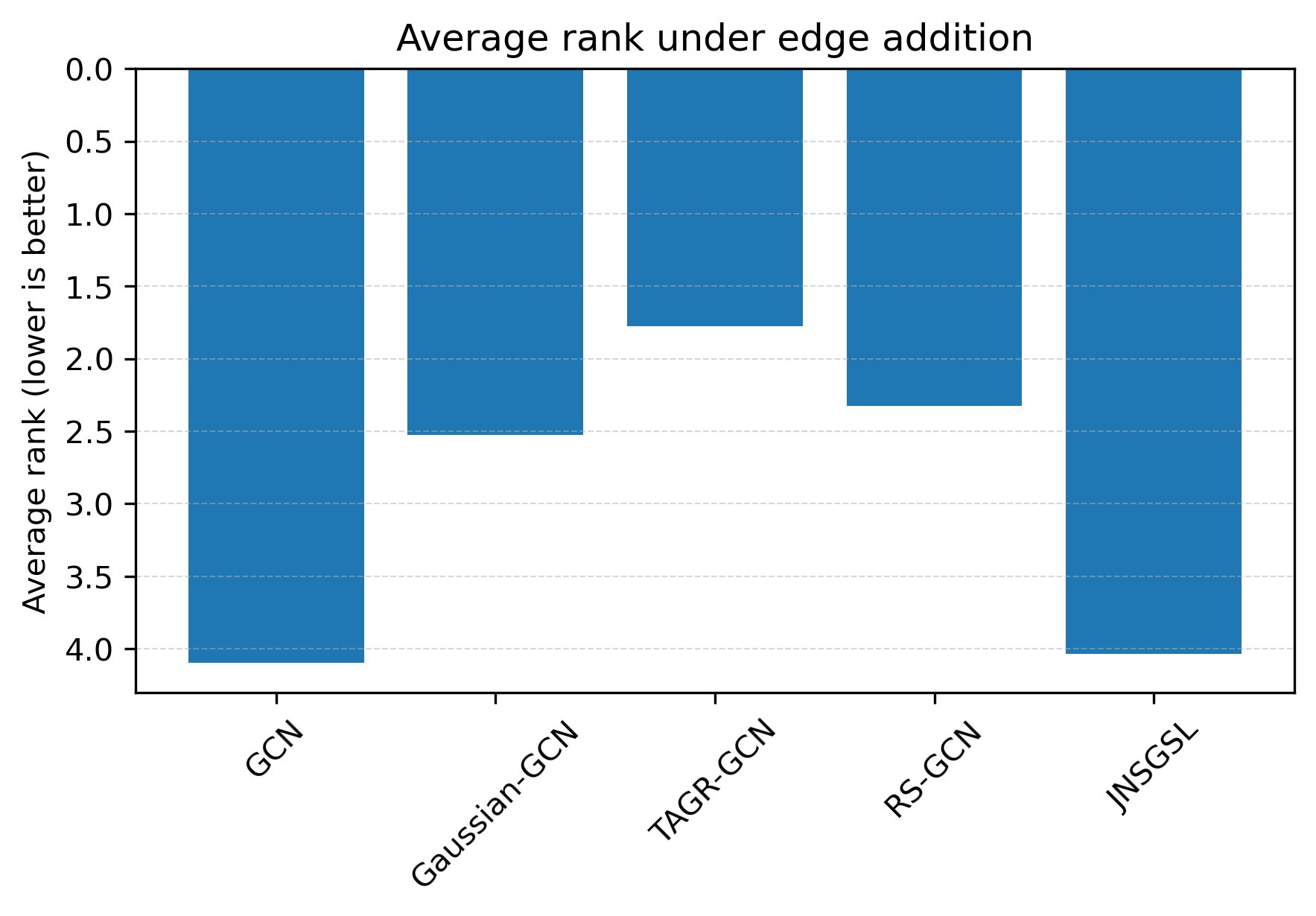}
        \label{fig:rank_add}}
    \quad
    \subfigure[Edge deletion.]{%
        \includegraphics[scale=.50]{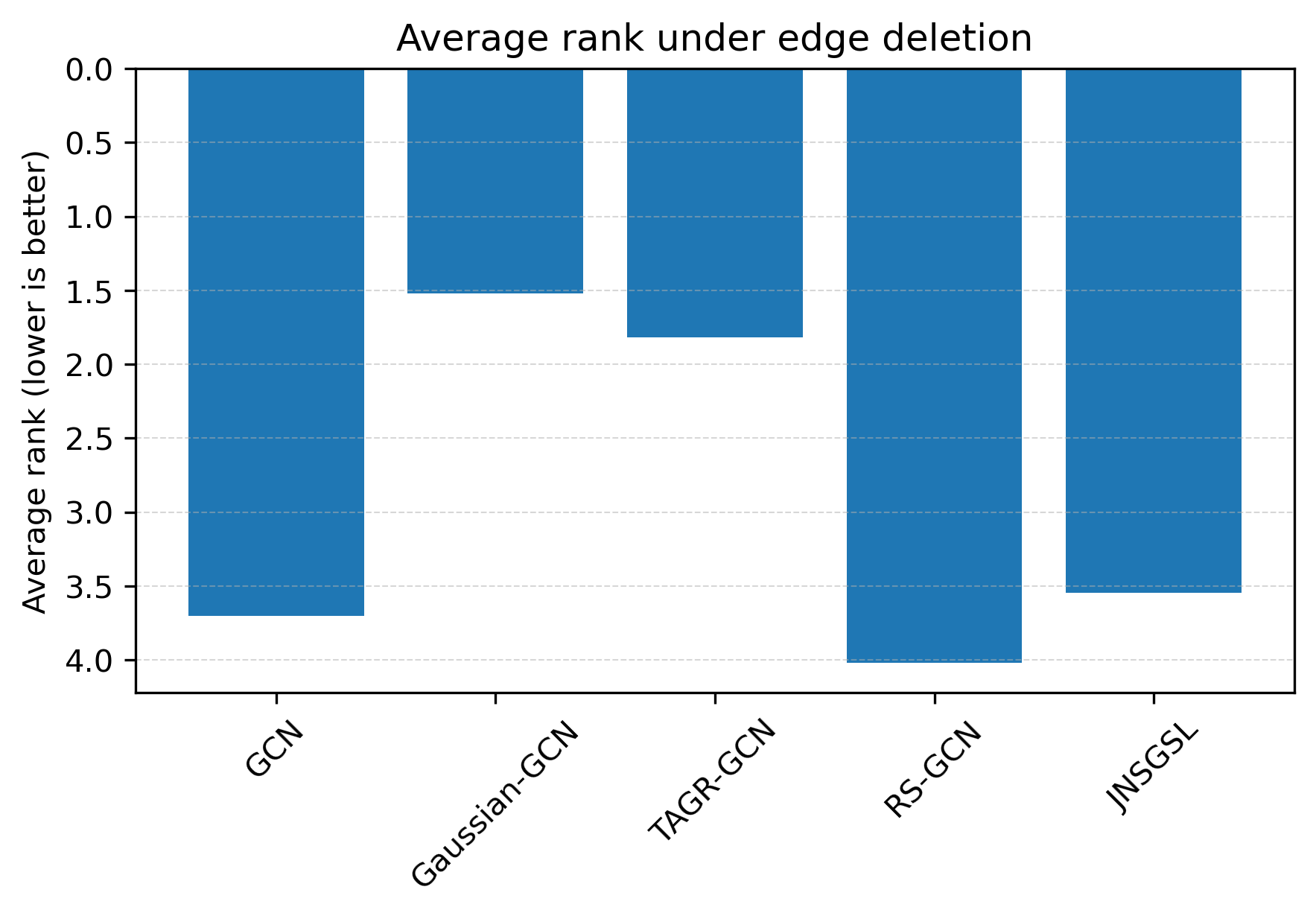}
        \label{fig:rank_del}}

\end{center}

\caption{Average ranking of methods across robustness curves. For each dataset and perturbation ratio, methods are ranked according to mean test accuracy, where a lower rank is better. TAGR-GCN obtains the best average rank under both edge addition and edge deletion, showing that although RS-GCN is strongest in some high-addition regimes and Gaussian-GCN is competitive in several deletion regimes, TAGR-GCN provides the most consistent overall robustness across perturbation types and datasets.}
\label{fig:average_rank}

\end{figure*}
% \begin{figure*}[t]
%     \centering
%     \begin{subfigure}[t]{0.48\textwidth}
%         \centering
%         \includegraphics[width=\linewidth]{average_rank_add.png}
%         \caption{Edge addition}
%         \label{fig:rank_add}
%     \end{subfigure}
%     \hfill
%     \begin{subfigure}[t]{0.48\textwidth}
%         \centering
%         \includegraphics[width=\linewidth]{average_rank_del.png}
%         \caption{Edge deletion}
%         \label{fig:rank_del}
%     \end{subfigure}

%     \caption{Average ranking of methods across robustness curves. For each dataset and perturbation ratio, methods are ranked according to mean test accuracy, where a lower rank is better. TAGR-GCN obtains the best average rank under both edge addition and edge deletion, showing that although RS-GCN is strongest in some high-addition regimes and Gaussian-GCN is competitive in several deletion regimes, TAGR-GCN provides the most consistent overall robustness across perturbation types and datasets.}
%     \label{fig:average_rank}
% \end{figure*}
To summarize performance across the entire perturbation range, Figure~\ref{fig:average_rank} reports the average rank of each method over all available datasets and perturbation ratios, where a lower rank indicates better average performance. TAGR-GCN obtains the best average rank under both edge addition and edge deletion. This result is important because TAGR is not always the top-performing method at every individual perturbation level. Instead, its strength lies in consistent performance across corruption regimes. RS-GCN is strongest in several severe edge-addition settings, while Gaussian-GCN is highly competitive under deletion noise, but TAGR-GCN provides the most balanced robustness across both types of structural perturbation.

These curves also illustrate the computational motivation behind TAGR. The method constructs a sparse repaired graph using feature-neighborhood search and local topology-aware reweighting, and then trains an ordinary GNN on the repaired graph. In contrast, graph structure learning methods may require dense pairwise scoring, additional graph-learning modules, or repeated optimization of graph parameters. Such costs become more significant on larger graphs and under robustness-curve evaluation, where each method must be run repeatedly across perturbation ratios and random seeds. TAGR is therefore positioned as a lightweight graph repair method: it improves robustness through a controlled sparse modification of the message-passing graph while preserving compatibility with standard GNN backbones.

\subsection{Accuracy--Homophily Analysis}
\label{subsec:accuracy_homophily}

\begin{figure*}[t]
    \centering
    \includegraphics[width=\textwidth]{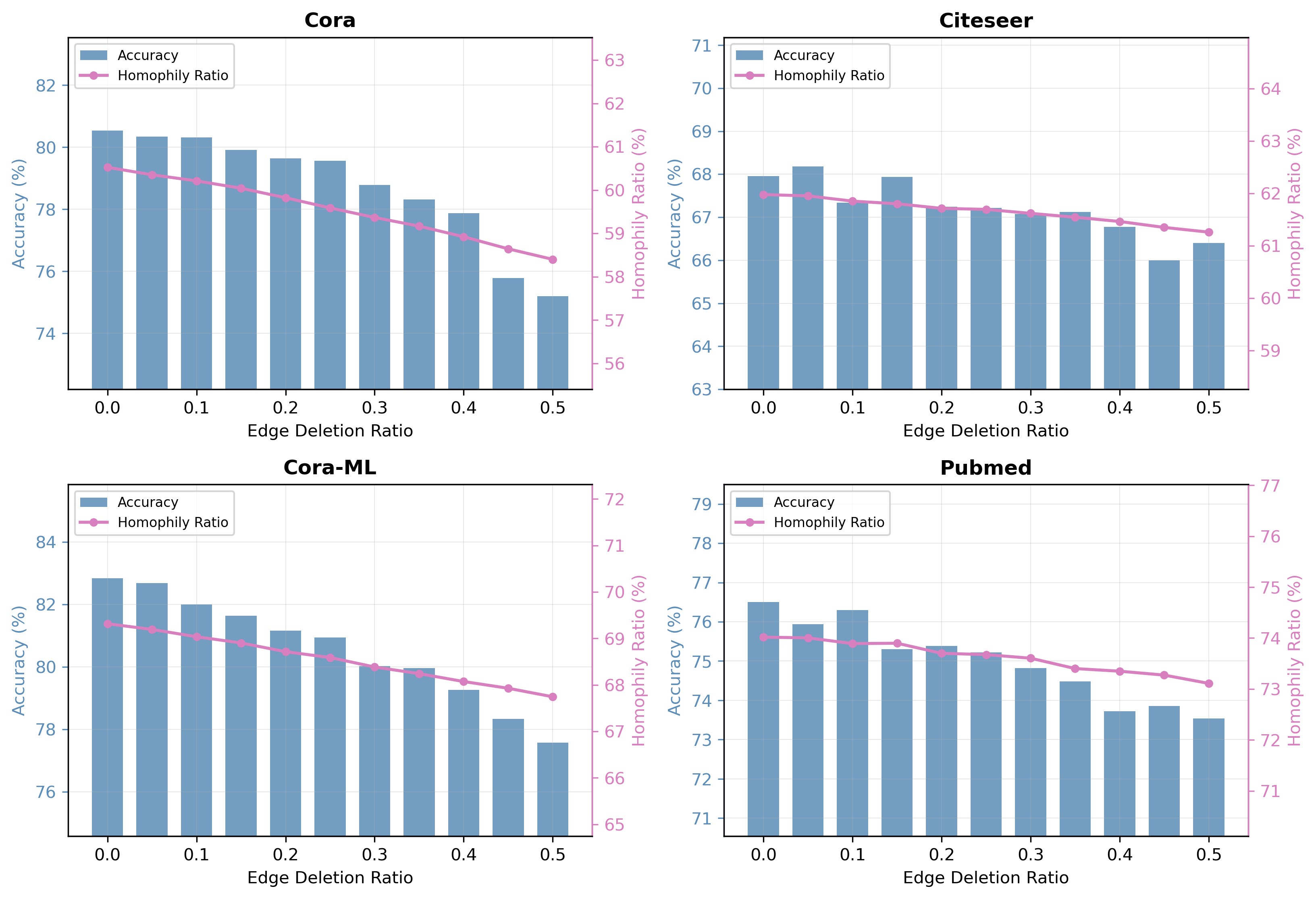}
    \caption{Accuracy--homophily analysis under edge deletion on Cora, Citeseer, Cora-ML, and Pubmed. Bars report the test accuracy of TAGR-GCN, and the line reports the homophily ratio of the TAGR-repaired graph. As observed edges are removed, accuracy decreases gradually while the repaired graph maintains relatively stable homophily, indicating that Gaussian feature-neighborhood repair helps compensate for missing propagation paths and preserves label-consistent message passing.}
    \label{fig:tagr_acc_homophily_del}
\end{figure*}

\begin{figure*}[t]
    \centering
    \includegraphics[width=\textwidth]{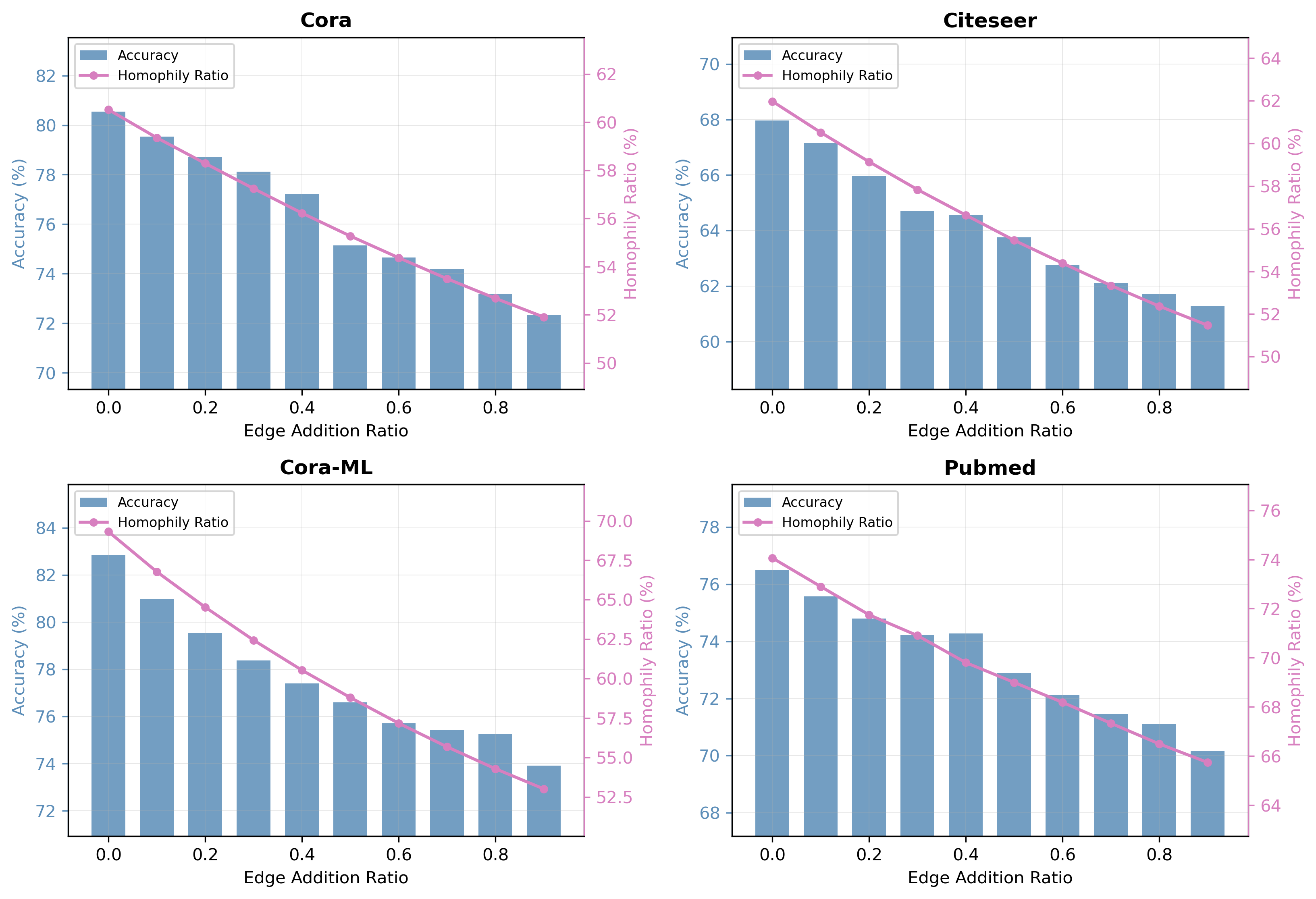}
    \caption{Accuracy--homophily analysis under edge addition on Cora, Citeseer, Cora-ML, and Pubmed. Bars report the test accuracy of TAGR-GCN, and the line reports the homophily ratio of the TAGR-repaired graph. As random edges are added, both homophily and accuracy decline, showing that spurious inter-class edges reduce the reliability of message passing even after graph repair.}
    \label{fig:tagr_acc_homophily_add}
\end{figure*}

To better understand how TAGR behaves under structural perturbations, we analyze the relationship between classification accuracy and the homophily ratio of the repaired graph. Homophily ratio is defined as the proportion of edges connecting nodes with the same class label. Although labels are not used by TAGR to construct the repaired graph, this quantity provides a useful diagnostic of whether the resulting message-passing topology is label-consistent. In Figs.~\ref{fig:tagr_acc_homophily_del} and~\ref{fig:tagr_acc_homophily_add}, the bars show the test accuracy of TAGR-GCN, while the line shows the homophily ratio of the TAGR-repaired graph.

Figure~\ref{fig:tagr_acc_homophily_del} shows the results under edge deletion. As more observed edges are removed, the graph becomes increasingly incomplete and useful propagation paths are lost. Consequently, TAGR-GCN accuracy gradually decreases on all datasets. However, the degradation is relatively smooth, especially under small and moderate deletion ratios. This indicates that the Gaussian feature-neighborhood repair component can compensate for missing observed edges by introducing auxiliary connections between feature-consistent nodes.

The homophily curves provide further insight. Under edge deletion, the homophily ratio of the repaired graph changes gradually rather than collapsing. This suggests that TAGR preserves a label-consistent message-passing structure even when the observed topology becomes sparse. On Cora, Cora-ML, and Pubmed, the homophily ratio remains within a relatively narrow range as deletion increases, and the corresponding accuracy decreases smoothly. Citeseer shows more variation in accuracy, but the repaired graph still avoids a sharp loss of homophilic structure. These trends support the interpretation that TAGR is effective under missing-edge perturbations because feature-neighborhood repair restores part of the lost connectivity.

Figure~\ref{fig:tagr_acc_homophily_add} shows a different behavior under edge addition. Randomly inserted edges can connect nodes from different classes, reducing the label consistency of the message-passing graph. As the addition ratio increases, both the homophily ratio and TAGR-GCN accuracy decrease consistently across all datasets. This confirms that spurious edges are particularly harmful because they do not merely remove useful paths; they actively introduce misleading propagation channels.

Despite this degradation, TAGR-GCN remains stable across the addition range compared with the severity of the perturbation. The Gaussian feature-neighborhood repair graph provides feature-consistent auxiliary neighborhoods, while the topology-aware residual reweighting reduces over-reliance on unreliable observed edges. Therefore, the repaired graph partially offsets the effect of noisy edge insertion, even though severe random addition remains more challenging than deletion.

The contrast between the two perturbation types is informative. Under deletion, homophily decreases slowly and accuracy degrades gradually, suggesting that TAGR can recover useful neighborhoods when observed edges are missing. Under addition, homophily drops more sharply and accuracy follows the same trend, indicating that randomly added edges directly reduce the quality of message passing. This analysis is consistent with the main robustness results: TAGR is particularly effective at repairing incomplete topology, while severe edge-addition noise remains a harder setting because it injects misleading structural signals. Overall, the accuracy--homophily analysis shows that TAGR improves robustness by maintaining a more reliable repaired graph under both missing-edge and noisy-edge perturbations.

\section{Discussion}
\label{sec:discussion}

The experimental results support the central premise of this work: the graph used for message passing should be treated as a repairable computational structure rather than as a fixed and fully reliable input. Standard GNNs define neighborhoods directly from the observed adjacency matrix. When the graph contains spurious edges, message passing may aggregate information from irrelevant nodes; when the graph is incomplete, useful information cannot propagate between related nodes. TAGR addresses both failure modes by constructing a sparse repaired graph before GNN training.

The results show that Gaussian feature-neighborhood repair is the main source of robustness. This indicates that node features contain relational information that is not fully captured by the observed topology. In attributed citation graphs, for example, two documents may be topically similar even when no citation edge connects them. By adding sparse feature-consistent auxiliary edges, TAGR restores part of this missing semantic connectivity, which explains its strong behavior under edge-deletion perturbations. The topology-aware residual plays a complementary role: it preserves the observed graph but reweights existing edges according to local feature and structural evidence. Thus, TAGR does not assume that the original graph is entirely wrong; it treats the topology as informative but imperfect.

The comparison with learned graph repair and graph structure learning methods clarifies the practical role of TAGR. Learned repair methods can be highly effective, especially under severe edge-addition noise, because they can adaptively suppress spurious connectivity. However, they often introduce additional graph-learning modules, dense pairwise scoring, or more complex optimization. TAGR instead uses a deterministic sparse repair operator. It remains competitive across both edge-addition and edge-deletion regimes while avoiding dense adjacency learning and preserving compatibility with standard GNN backbones.

The backbone and homophily analyses provide a consistent explanation of the results. The largest gains appear for GCN because its propagation is directly controlled by the repaired adjacency matrix. GAT and GraphSAGE also benefit, but the effect is moderated by their aggregation mechanisms. The accuracy--homophily analysis further shows that TAGR maintains relatively stable label-consistent structure under deletion, whereas severe edge addition remains harder because it injects misleading inter-class edges.

TAGR also has limitations. Its Gaussian repair component depends on the quality of node features, and its residual reweighting uses fixed local statistics rather than a learned task-specific scoring function. Moreover, the repaired graph is constructed before training and remains fixed during optimization. These choices make TAGR simple, interpretable, and efficient, but they may limit adaptivity in graphs with noisy features or task-specific structural patterns. Future work will explore adaptive and task-aware variants of TAGR, robustness under feature noise, and extensions to inductive, heterogeneous, temporal, and large-scale graph learning settings.

\section{Conclusion}
\label{sec:conclusion}

We proposed TAGR, a topology-aware Gaussian graph repair framework for improving the robustness of GNN message passing under structurally imperfect graphs. TAGR is motivated by the observation that real graphs may contain both spurious and missing edges. Instead of only removing unreliable edges or learning a dense task-dependent graph structure, TAGR constructs a sparse repaired graph using two complementary mechanisms: adaptive Gaussian feature-neighborhood repair, which restores feature-consistent propagation paths, and topology-aware residual reweighting, which refines observed edges using local feature and structural evidence.

The repaired graph can be used directly with standard message-passing architectures without modifying their neural components or introducing an additional graph generator. Experiments on benchmark citation networks under clean, edge-addition, and edge-deletion settings show that TAGR improves robustness over base GNNs and remains competitive with stronger learned graph repair and graph structure learning baselines. Robustness curves further demonstrate that TAGR-GCN provides stable performance across perturbation ratios, supporting its effectiveness under both noisy and incomplete topology.

The experimental analysis shows that Gaussian feature-neighborhood repair is the main source of robustness, while topology-aware residual reweighting provides additional stabilization by preserving and refining useful observed structure. Overall, the results suggest that robust graph learning does not always require dense graph structure learning or complex graph-generation modules. Sparse, interpretable, and topology-aware graph repair can provide a reliable message-passing substrate while remaining compatible with existing GNN backbones.

Future work will explore adaptive and task-aware variants of TAGR, robustness under feature noise, and extensions to inductive, heterogeneous, temporal, and large-scale graph learning settings.
\section{Appendix-I}

\subsection{Additional Analysis on Flickr and Cornell}
\label{subsec:appendix_flickr_cornell}

Table~\ref{tab:appendix_dataset_stats} summarizes Flickr and Cornell, which are included as diagnostic datasets for analyzing the topology-aware residual. These datasets are not used to define the main benchmark comparison; rather, they provide complementary graph regimes for mechanism analysis. Flickr is substantially larger and denser, whereas Cornell is a small WebKB graph with far fewer nodes and edges. This contrast allows us to examine whether the residual component behaves similarly across graph scales and structural regimes.

\begin{table}[htbp!]
\centering
\caption{
Summary statistics for the Flickr and Cornell datasets.
}
\label{tab:appendix_dataset_stats}

\begin{tabular}{lrrrrrrr}
\toprule
Dataset & Classes & Nodes & Edges & Features & Train & Val & Test \\
\midrule
Flickr  & 9 & 7,575 & 239,738 & 12,047 & 757 & 1,515 & 5,303 \\
Cornell & 5 & 183   & 295     & 1,703  & 88  & 58    & 37 \\
\bottomrule
\end{tabular}
\end{table}
Table~\ref{tab:appendix_flickr_cornell} reports five-seed clean-setting results on these datasets. TAGR-GCN obtains the highest mean test accuracy on both datasets. The comparison with Gaussian-GCN shows that the residual component has different effects across the two datasets. On Flickr, the improvement is moderate, suggesting that the residual mainly provides an additional refinement. On Cornell, the improvement is much larger, suggesting that topology-aware reweighting can become an active repair mechanism when local structural cues are informative.

\begin{table}[htbp!]
\centering
\caption{
Five-seed clean-setting results on Flickr and Cornell. We report mean test accuracy and standard deviation. The best result for each dataset is shown in bold.
}
\label{tab:appendix_flickr_cornell}
\begin{tabular}{llc}
\toprule
Dataset & Method & Test Accuracy \\
\midrule
\multirow{5}{*}{Flickr}
& TAGR-GCN & $\mathbf{0.8797}\pm0.018$ \\
& Gaussian-GCN & $0.8679\pm0.0020$ \\
& RS-GCN & $0.8567\pm0.0038$ \\
& GCN & $0.6382\pm0.0085$ \\
& JNSGSL & $0.6254\pm0.0103$ \\
\midrule
\multirow{5}{*}{Cornell}
& TAGR-GCN & $\mathbf{0.8203}\pm0.0581$ \\
& RS-GCN & $0.8054\pm0.0884$ \\
& JNSGSL & $0.5676\pm0.0936$ \\
& Gaussian-GCN & $0.5514\pm0.2116$ \\
& GCN & $0.2270\pm0.0308$ \\
\bottomrule
\end{tabular}
\end{table}

\subsection{Is the Topology-Aware Residual Only a Stabilizer?}
\label{subsec:residual_stabilizer}

TAGR combines two complementary repair mechanisms: adaptive Gaussian feature-neighborhood repair and topology-aware residual reweighting. The Gaussian component adds missing feature-consistent edges, whereas the topology-aware residual reweights observed edges according to local topology-feature consistency. Since the residual component introduces additional design choices, it is important to determine whether it merely stabilizes Gaussian repair or whether it can act as an independent repair signal.

\begin{figure}[htbp!]
    \centering
    \includegraphics[width=0.55\linewidth]{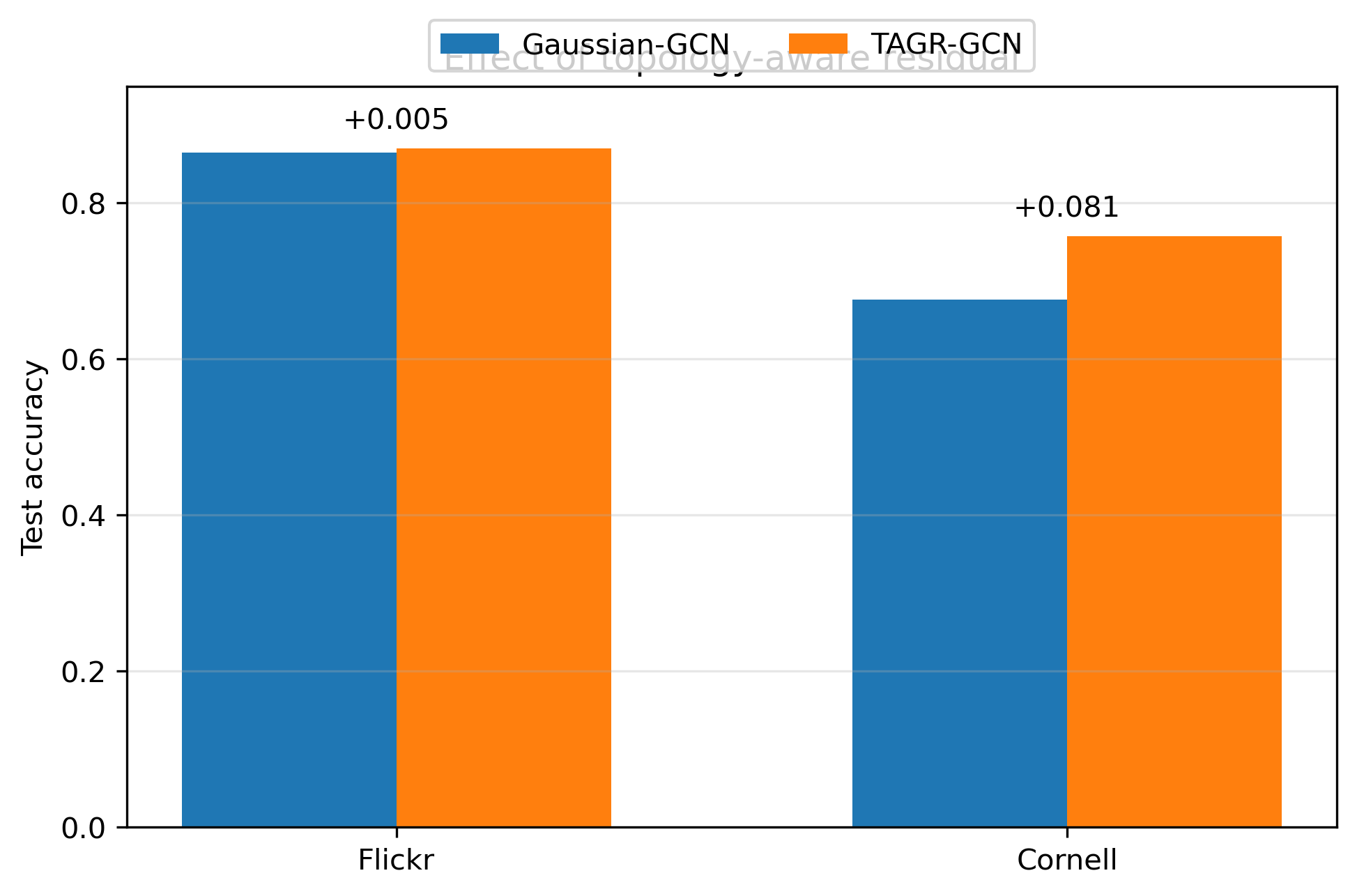}
    \caption{
    Test accuracy comparison between Gaussian-GCN and the best TAGR-GCN configuration on Flickr and Cornell under the clean seed-1234 setting. Gaussian-GCN corresponds to $\lambda=0$, while TAGR-GCN uses $\lambda>0$. TAGR-GCN provides a moderate gain on Flickr and a much larger gain on Cornell, indicating that the topology-aware residual has a dataset-dependent effect.
    }
    \label{fig:extra_residual_bar}
\end{figure}
To address this question, we perform a sensitivity analysis on Flickr and Cornell under the clean setting. For both datasets, we fix the GCN training hyperparameters and the Gaussian repair hyperparameters, and vary only the topology-aware residual. Specifically, we vary the residual strength $\lambda$, the residual-score profile, and the clipping range of the residual multiplier. The setting $\lambda=0$ corresponds to Gaussian-GCN, while $\lambda>0$ corresponds to TAGR-GCN.

Figure~\ref{fig:extra_residual_bar} compares Gaussian-GCN with the best TAGR-GCN configuration in terms of test accuracy. On Flickr, TAGR-GCN improves over Gaussian-GCN from $0.864$ to $0.869$ in the seed-1234 sensitivity setting. The five-seed comparison in Table~\ref{tab:appendix_flickr_cornell} confirms this trend, where TAGR-GCN improves the mean accuracy from $0.8679$ to $0.8797$. This indicates that topology-aware reweighting provides a stable but moderate gain on Flickr.

\begin{figure}[htbp!]
    \centering
    \includegraphics[width=1.1\linewidth]{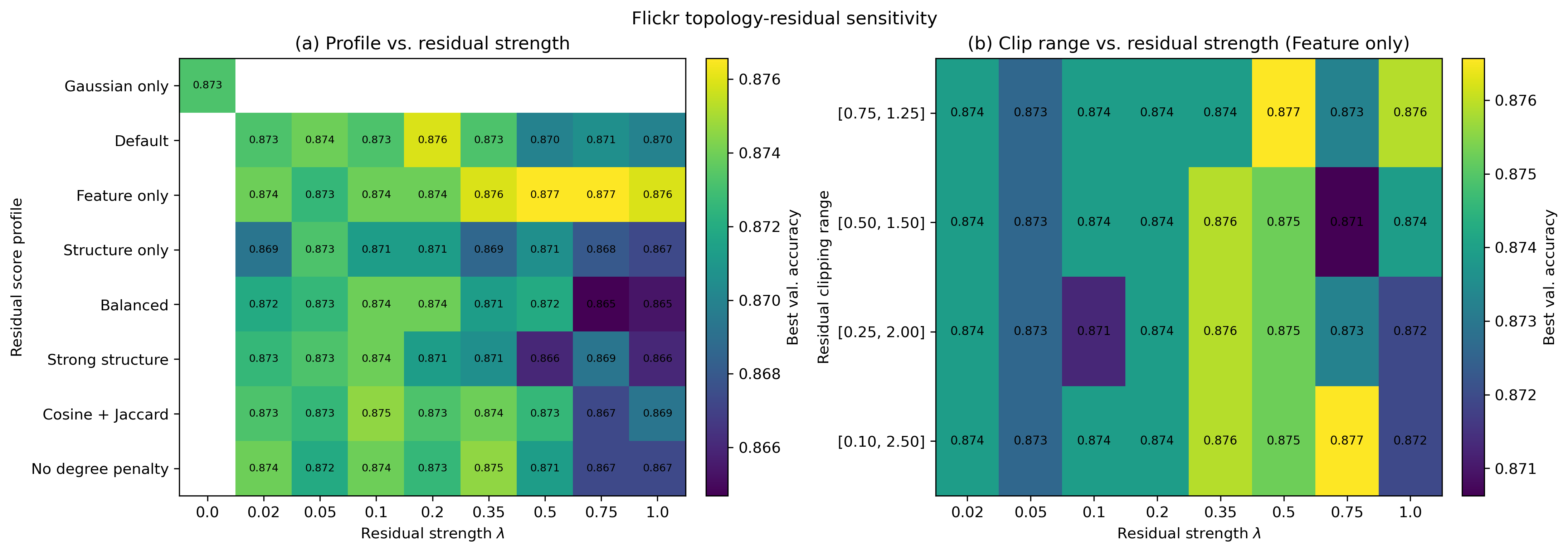}
    \caption{
    Topology-aware residual sensitivity on Flickr. Panel (a) shows validation accuracy as a function of residual-score profile and residual strength. Panel (b) shows validation accuracy as a function of clipping range and residual strength for the best residual profile. The broad high-performing region indicates that Flickr is relatively insensitive to the exact residual design.
    }
    \label{fig:flickr_combined_sensitivity}
\end{figure}

Figure~\ref{fig:flickr_combined_sensitivity} analyzes Flickr in more detail. Panel (a) shows the topology-profile heatmap, where validation accuracy is reported as a function of residual-score profile and residual strength. The high-performing region is relatively broad across feature-dominant residual profiles and moderate residual strengths. Panel (b) fixes the best residual profile and varies the clipping range and residual strength. The resulting heatmap shows only small variations across clipping ranges. These results suggest that Flickr represents a feature-dominant regime: Gaussian feature-neighborhood repair already captures most of the useful signal, and the topology-aware residual mainly provides a stabilizing refinement.

\begin{figure}[htbp!]
    \centering
    \includegraphics[width=1.1\linewidth]{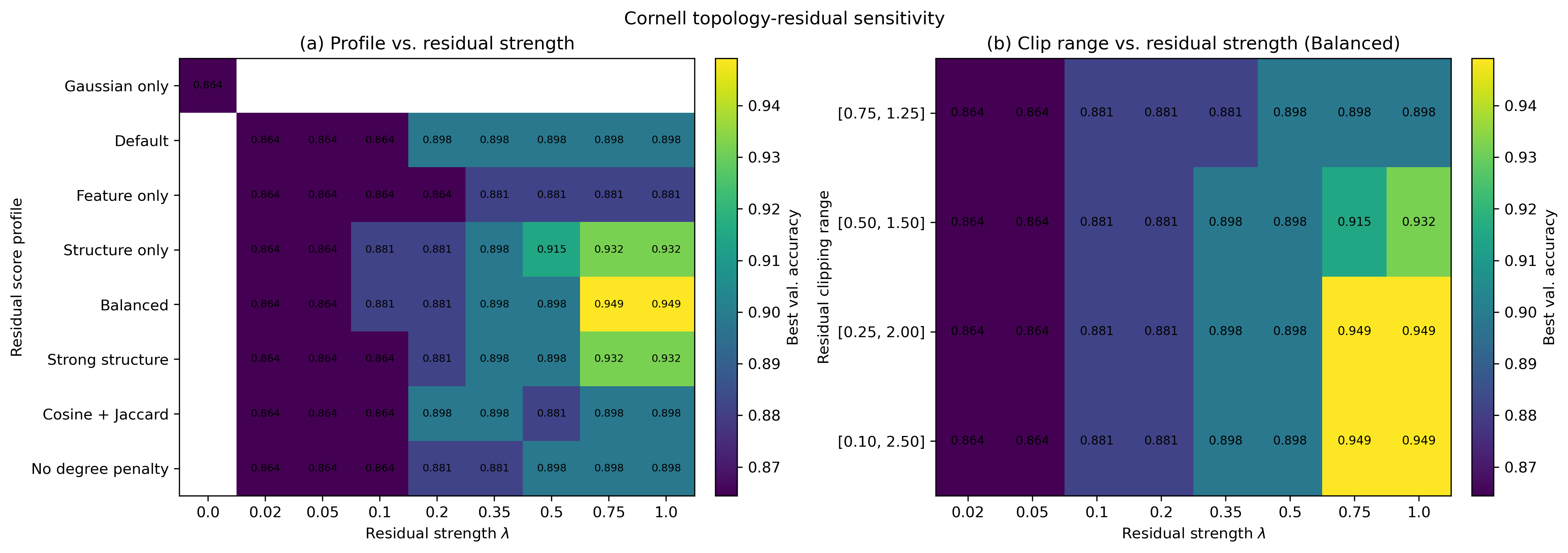}
    \caption{
    Topology-aware residual sensitivity on Cornell. Panel (a) shows validation accuracy as a function of residual-score profile and residual strength. Panel (b) shows validation accuracy as a function of clipping range and residual strength for the best residual profile. Cornell benefits from stronger residual strengths and wider clipping ranges, indicating that topology-aware reweighting acts as an active graph-repair signal.
    }
    \label{fig:cornell_combined_sensitivity}
\end{figure}

Cornell exhibits a different pattern. In Table~\ref{tab:appendix_flickr_cornell}, TAGR-GCN improves over Gaussian-GCN from $0.5514$ to $0.8203$ and also exceeds RS-GCN in mean accuracy. Figure~\ref{fig:cornell_combined_sensitivity} shows that balanced and structure-aware residual profiles achieve substantially higher validation accuracy as the residual strength increases. In particular, Panel (a) shows a strong dependence on the residual-score profile and $\lambda$, while Panel (b) shows that wider clipping ranges and larger $\lambda$ values are most effective for the balanced profile. These results indicate that Cornell is a structure-sensitive regime: reweighting observed edges has a large effect on GCN propagation, and the topology-aware residual becomes an active graph-repair signal rather than a small stabilizer.

These findings indicate that the topology-aware residual is not redundant, but its effect depends on the graph regime. On feature-dominant graphs such as Flickr, it stabilizes and moderately improves Gaussian repair. On structure-sensitive graphs such as Cornell, it becomes a decisive repair mechanism that substantially improves over Gaussian-only repair. This distinction explains why TAGR can remain close to Gaussian-GCN on some datasets while producing large gains on others.
\bibliographystyle{plainnat}
\bibliography{bib}
\end{document}